\documentclass[pdflatex,sn-mathphys-num]{sn-jnl}

\usepackage{graphicx}%
\usepackage{multirow}%
\usepackage{amsmath,amssymb,amsfonts}%
\usepackage{amsthm}%
\usepackage{mathrsfs}%
\usepackage[title]{appendix}%
\usepackage{xcolor}%
\usepackage{textcomp}%
\usepackage{manyfoot}%
\usepackage{booktabs}%
\usepackage{algorithm}%
\usepackage{algorithmicx}%
\usepackage{algpseudocode}%
\usepackage{listings}%
\usepackage{subcaption}
\usepackage{placeins} 
\usepackage{caption}
\theoremstyle{thmstyleone}%
\usepackage{adjustbox}

\theoremstyle{thmstyletwo}%

\theoremstyle{thmstylethree}%

\begin{document}

\title[Multi-task Neural Diffusion Processes]{Multi-task Neural Diffusion Processes}

\author[1]{\fnm{Joseph} \sur{Rawson}}

\equalcont{These authors contributed equally to this work.}

\author*[1]{\fnm{Domniki} \sur{Ladopoulou}}\email{domna.ladopoulou.22@ucl.ac.uk}
\equalcont{These authors contributed equally to this work.}

\author[1,2]{\fnm{Petros} \sur{Dellaportas}}\email{p.dellaportas@ucl.ac.uk}

\affil[1]{\orgdiv{Department of Statistical Science}, \orgname{University College London}, \orgaddress{\street{Gower Street}, \city{London}, \postcode{WC1E 6BT}, \country{United Kingdom}}}

\affil[2]{\orgdiv{Department of Statistics}, \orgname{Athens University of Economics and Business}, \orgaddress{\street{Patission 76}, \city{Athens}, \postcode{104 34}, \country{Greece}}}

\abstract{Neural diffusion processes provide a scalable, non-Gaussian approach to modelling distributions over functions, but existing formulations are limited to single-task inference and do not capture dependencies across related tasks. We propose multi-task neural diffusion processes (MT-NDPs), an extension that incorporates a task encoder to enable task-conditioned probabilistic regression and few-shot adaptation across related functions. The task encoder extracts a low-dimensional representation from context observations and conditions the diffusion model on this representation, allowing information sharing across tasks while accommodating variable-sized context sets and preserving permutation invariance with respect to the ordering of context observations. We evaluate MT-NDPs through synthetic experiments and two real-world applications involving wind turbine data and school examination outcomes. The results demonstrate improved predictive accuracy relative to single-task diffusion baselines when context observations are available, alongside competitive performance against neural-process and Gaussian-process alternatives. In the wind-energy application, MT-NDPs achieve the lowest average coverage error among the evaluated models at larger context sizes, demonstrating improved predictive uncertainty calibration. Overall, our findings highlight the potential of task-conditioned diffusion models for uncertainty-aware multi-task regression and few-shot adaptation to previously unseen tasks.}


\maketitle
\section{Introduction}

Diffusion probabilistic models have emerged as a powerful approach for learning complex distributions via iterative denoising \cite{sohldickstein2015deepunsupervisedlearningusing, ho2020denoisingdiffusionprobabilisticmodels}. Neural diffusion processes (NDPs) are a class of probabilistic models that learn distributions over functions by extending diffusion-based generative modelling to function space \cite{dutordoir2023neural}. They operate by iteratively corrupting and denoising function values, avoiding explicit Gaussian assumptions while retaining scalability through neural network (NN) parameterisations. In the regression setting, NDPs support uncertainty-aware prediction and allow conditioning on context observations at inference time via conditional inference \cite{dutordoir2023neural}. These properties make NDPs a flexible foundation for probabilistic regression in complex and non-linear settings.

We introduce multi-task neural diffusion processes (MT-NDPs) for regression across related tasks that share common structure while exhibiting task-specific variation, such as multi-sensor measurements, environmental data, and wind turbines operating within a wind farm \cite{alvarez2012kernels, wang2016locally, Fiocchi_2025}. Joint modelling can improve data efficiency and predictive performance in low-data regimes \cite{caruana1997multitask}, whereas existing NDP formulations lack explicit mechanisms for learning task-specific representations across related functions. 
Prior work on multi-task probabilistic regression has explored several approaches for sharing information across different tasks. 
Multi-task Gaussian processes (GPs) provide a Bayesian framework for joint inference via structured cross-task covariance models, including autoregressive formulations that handle incomplete observations across tasks \cite{bonilla2007multi, alvarez2012kernels, requeima2019gaussian}. While effective, these methods scale poorly to large datasets and depend strongly on kernel and covariance design \cite{Rasmussen2006GaussianLearning}.  Neural alternatives such as neural processes (NPs) improve scalability by learning task representations from context sets, and multi-task extensions enable transfer learning across related functions \cite{garnelo2018neuralprocesses, kim2019attentiveneuralprocesses, kim2022multitaskneuralprocesses}. 
However, these models often suffer from underfitting and inconsistent uncertainty calibration. 
NDPs address these limitations by learning expressive, non-Gaussian distributions over functions \cite{dutordoir2023neural}, but existing formulations are restricted to single-task inference. 

Wind energy systems provide a representative and practically relevant example of multi-task regression, where multiple wind turbines operate under similar environmental conditions while exhibiting turbine-specific behaviour. Accurate and uncertainty-aware wind power prediction is essential for grid stability and maintenance planning, particularly given that operation and maintenance costs account for a substantial fraction of total wind farm expenditure \cite{en13123132, Hadjoudj2023AOpportunities}. Prior work has explored a range of probabilistic and machine learning approaches for wind power prediction, including GPs \cite{2026probabilistic, rogers2020, pandit2020, chen2014}, quantile regression \cite{yu2019probabilistic,yu2021,zhou2022}, mixture density networks \cite{zhang2020}, and NN–based probabilistic models \cite{Fiocchi_2025}. While these methods provide varying degrees of flexibility and uncertainty quantification, many either scale poorly to large datasets or lack principled mechanisms for sharing information across turbines.

To address these challenges, we propose MT-NDPs, which incorporate a task encoder that maps context observations to a low-dimensional task representation and conditions the diffusion denoiser on this representation. This design enables information sharing across related tasks while preserving the set-structured formulation and probabilistic flexibility of NDPs. As a result, MT-NDPs support few-shot adaptation to unseen tasks by conditioning on limited task-specific context at inference time. 

 We evaluate the proposed framework through controlled synthetic experiments and two real-world regression applications. First, we consider wind power prediction using supervisory control and data acquisition (SCADA) data, examining the ability of MT-NDPs to model complex turbine-specific behaviour and adapt to unseen wind turbines under varying context sizes. Second, synthetic experiments investigate the influence of embedding dimensionality and the effect of context size on generalisation to unseen tasks. Finally, we evaluate MT-NDPs on the Exam dataset \cite{goldstein1993multilevel}, providing an established multi-task regression benchmark to assess predictive accuracy, uncertainty calibration, and generalisation to previously unseen tasks. We further investigate how architectural choices, including the task encoder and the dimensionality of its representation, influence predictive performance and uncertainty quantification. Across the real-world applications, we compare MT-NDPs against relevant neural-process and Gaussian-process baselines. The results demonstrate that MT-NDPs achieve competitive predictive accuracy and improved uncertainty calibration in several experimental settings, highlighting the benefits of task-conditioned diffusion models for multi-task regression.

The remainder of the paper is organised as follows. Section~\ref{sec:ndp} introduces NDPs and their core properties. Section~\ref{sec:mtndp} presents the proposed multi-task extension. Section~\ref{sec:experimental_design} describes the datasets and experimental setups, and Section~\ref{sec:results} reports the empirical results. Section~\ref{sec:discussion} discusses implications and limitations, and Section~\ref{sec:conclusion} concludes.

\section{Background: Neural Diffusion Processes} \label{sec:ndp}

This section reviews the key
concepts used throughout the manuscript.  Diffusion probabilistic models learn complex distributions by gradually corrupting data with noise and training an NN  to reverse this process \cite{ho2020denoisingdiffusionprobabilisticmodels}. NDPs adapt this framework from finite-dimensional data vectors to function values, enabling probabilistic regression with uncertainty quantification \cite{dutordoir2023neural}. For completeness, we summarise the diffusion framework in Appendix~\ref{subsec:dpm}. By operating directly in function space, NDPs avoid the restrictive Gaussian assumptions of GPs while retaining scalability through neural architectures. Compared to NPs, NDPs yield better-calibrated predictive distributions and a more consistent treatment of context and target points. On synthetic benchmarks, NDPs have been shown to closely emulate GP behaviour while scaling more effectively to larger datasets \cite{dutordoir2023neural}. 

The core idea of NDPs is to replace the fixed finite-dimensional data vector used in standard diffusion models with function evaluations $(X, y)$. Here $X \in \mathbb{R}^{N \times D}$ is the matrix of input locations and $y \in \mathbb{R}^N$ the corresponding outputs. In the forward process, Gaussian noise is gradually added to the function values $y$, while the inputs $X$ remain fixed. After $T$ steps, the corrupted outputs resemble white noise. The reverse process is parameterised by an NN that predicts the noise injected at each step, thereby enabling recovery of the original function values.

We now formalise this intuition by introducing the training data representation used by NDPs. Each training instance corresponds to a sampled function $f_i:\mathbb{R}^D\to \mathbb{R}$ evaluated at a finite set of inputs $x_i \in \mathbb{R}^{N \times D}$ with outputs $y_i = f_i(x_i) \in \mathbb{R}^N$. The overall training dataset is therefore
\begin{equation}
    \mathcal{D} = \big\{ (x_i, y_i) \big\}_{i=1}^K,
    \qquad
    x_i \in \mathbb{R}^{N \times D},\;\;
    y_i \in \mathbb{R}^N,
    \label{eq:ndp_dataset}
\end{equation}
where $K$ is the number of sampled functions, $N$ is the number of input locations per function, and $D$ is the input dimension.

  \vspace{0.5em}
\noindent\textbf{Forward process.}  Let $x_0 = x$ and $y_0 = y$. Then the NDPs
gradually add noise following
\begin{equation}
    q\!\left(
    \begin{bmatrix}
    x_t \\
    y_t
    \end{bmatrix}
    \Bigg|
    \begin{bmatrix}
    x_{t-1} \\
    y_{t-1}
    \end{bmatrix}
    \right)
    = \mathcal{N}\!\left(
    y_t ; \sqrt{1 - \beta_t}\, y_{t-1}, \, \beta_t I
    \right)
    \label{eq:ndp_f}
\end{equation} for each timestep $t$ where $\beta_t$ is the noise from the variance schedule \cite{dutordoir2023neural}. This forward process is directly analogous to the standard diffusion formulation (Eq.~\ref{eq:ddpm_f} in Appendix~\ref{subsec:dpm}), but adapted to function-space regression by explicitly separating input locations $x_t$ from function values $y_t$. This corresponds to adding Gaussian noise to the function
values $y_t$ while keeping the input locations $x_t$ fixed for all $t$. By the final timestep ($t = T$), the function values $y_t$ will look like Gaussian noise from $\mathcal{N}(0,1)$.

  \vspace{0.5em}

\noindent\textbf{Reverse process.}  The reverse process is parameterised by an NN that predicts the noise injected during the forward diffusion process \cite{ho2020denoisingdiffusionprobabilisticmodels}. The reverse transition is given by \cite{dutordoir2023neural}:

\begin{equation}
    p_\theta\!\left(
    \begin{bmatrix}
    x_{t-1} \\
    y_{t-1}
    \end{bmatrix}
    \Bigg|
    \begin{bmatrix}
    x_t \\
    y_t
    \end{bmatrix}
    \right)
    = \mathcal{N}\!\left(
    y_{t-1} ; \mu_\theta(x_t, y_t, t), \, \tilde{\beta}_t I
    \right)
    \label{eq:ndp_b}
\end{equation} with mean

\begin{equation}
    \mu_\theta(x_t, y_t, t)
    = \frac{1}{\sqrt{\alpha_t}}
    \left(
    y_t - \frac{\beta_t}{\sqrt{1 - \bar{\alpha}_t}} \,
    \epsilon_\theta(x_t, y_t, t)
    \right),
    \label{eq:ndp_mean1}
\end{equation} where the noise model $ \epsilon_\theta: \mathbb{R}^{N \times D} \times \mathbb{R}^N \times \mathbb{R} \;\to\; \mathbb{R}^N$ takes as input the non-corrupted $x_t$ as well as the corrupted $y_t$, and the timestep $t$. As in standard diffusion models, $\alpha_t = 1-\beta_t$ and $\bar{\alpha}_t = \prod_{j=1}^t \alpha_j$. 

 \vspace{0.5em}
\noindent\textbf{Training objective.}  Training proceeds by minimising a denoising score-matching objective,
\begin{equation}
    L_\theta = 
    \mathbb{E}_{t, x_0, y_0, \epsilon}
    \left[
    \left\|
    \epsilon - \epsilon_\theta(x_0, y_t, t)
    \right\|^2
    \right],
    \label{eq:ndp_loss}
\end{equation}
where $x_0 = x_t$ for all timesteps $t$, since the diffusion process is applied only to the outputs while the inputs remain fixed. The corrupted outputs are generated through the following reparameterisation:
\begin{equation}
    y_t = \sqrt{\bar{\alpha}_t}\, y_0 
    + \sqrt{1 - \bar{\alpha}_t}\, \epsilon,
    \qquad \epsilon \sim \mathcal{N}(0, I).\label{eq:y_t}
\end{equation}

\noindent\textbf{Unconditional sampling.} Given some input locations $x$, we can sample unconditionally from the trained model prior by starting with random noise and following the reverse process for $T$ timesteps. Specifically, we initialise $y_T \sim \mathcal{N}(0, I)$ and then apply the parametrised kernel for $t = T, \dots, 1$. At each step, the NN predicts the noise added in the forward process, and new noise is reintroduced. 

 \vspace{0.5em}
\noindent\textbf{Conditional sampling.} One of the main strengths of NDPs, as with NPs, is their ability to quickly adapt with only a small amount of data. Conditional sampling is achieved by reinjecting context points at each timestep, using a slight adaptation of the RePaint algorithm \cite{lugmayr2022RePaintinpaintingusingdenoising}. This guides the reverse process to remain consistent with the context set. 

Formally, we are interested in $p(y^{*}_{0} \mid x^{*}_{0}, D)$, where $D = (x^{c}_{0} \in \mathbb{R}^{M \times D},\ y^{c}_{0} \in \mathbb{R}^{M})$ is the set of context points. The target is initialised as $y^{*}_{T} \sim \mathcal{N}(0, I)$. At each timestep, we add noise to the context points, obtaining $y^{c}_{t}$ from the forward process. We then take the union of the noisy context points and the input locations $x_{0} = \{x^{*}_{0},\ x^{c}_{0}\}$, together with the union of the target outputs and context outputs $y_{t} = \{y^{*}_{t},\ y^{c}_{t}\}$. Finally, the reverse process kernel samples
\begin{equation}
    y_{t-1} \sim \mathcal{N}\!\left( 
        \frac{1}{\sqrt{\alpha_{t}}}
        \left( y_{t} - \frac{\beta_{t}}{\sqrt{1 - \bar{\alpha}_{t}}}\, 
        \epsilon_{\theta}(x_{0}, y_{t}, t) \right),
        \ \tilde{\beta}_{t} I
    \right).
    \label{eq:ndp_cond}
\end{equation} Repeating this for $t=T, T-1, \dots, 1$ means that, at each backward step, the context points steer the process in the desired direction. 

\subsection{Denoising Network Architecture}
\label{subsec:noise_model_arch}
The noise model architecture follows \cite{dutordoir2023neural}, designed to generate sensible prior distributions over functions. While any NN could in principle serve as the denoiser, two structural properties are particularly desirable: (i) \textit{Input-size agnosticity:} the model should work for arbitrary numbers of input points $N$ and input dimensions $D$, producing consistent priors whether inference is carried out at one or many locations. This is achieved by replicating the outputs $y_t \in \mathbb{R}^N$ across $D$ copies so that the network weights do not depend directly on $N$ or $D$. As a result, one trained NDP can be applied across datasets of varying size and dimensionality. (ii) \textit{Equivariance and invariance:} predictions should not depend on the ordering of inputs, context points, or feature dimensions. Reordering inputs should not change the probability of the data, just as kernels such as the radial basis function (RBF) in GPs are invariant to permutations.
To enforce the equivariance and invariance, \cite{dutordoir2023neural} proposed a bi-dimensional attention block, $A_t : \mathbb{R}^{N \times D \times H} \to \mathbb{R}^{N \times D \times H}$. This multi-head self-attention mechanism operates jointly across input locations and feature dimensions. Attention weights are computed via dot-product similarity, and multiple heads are run in parallel to capture complementary relationships. This design ensures permutation equivariance and invariance while retaining the flexibility of neural architectures, addressing limitations that are often overlooked in standard NN models.

\subsection{Illustrative Examples of Forward and Reverse Diffusion}
\label{subsec:illustrative_examples}

\begin{figure}[!ht]
\centering
\includegraphics[width=1\linewidth]{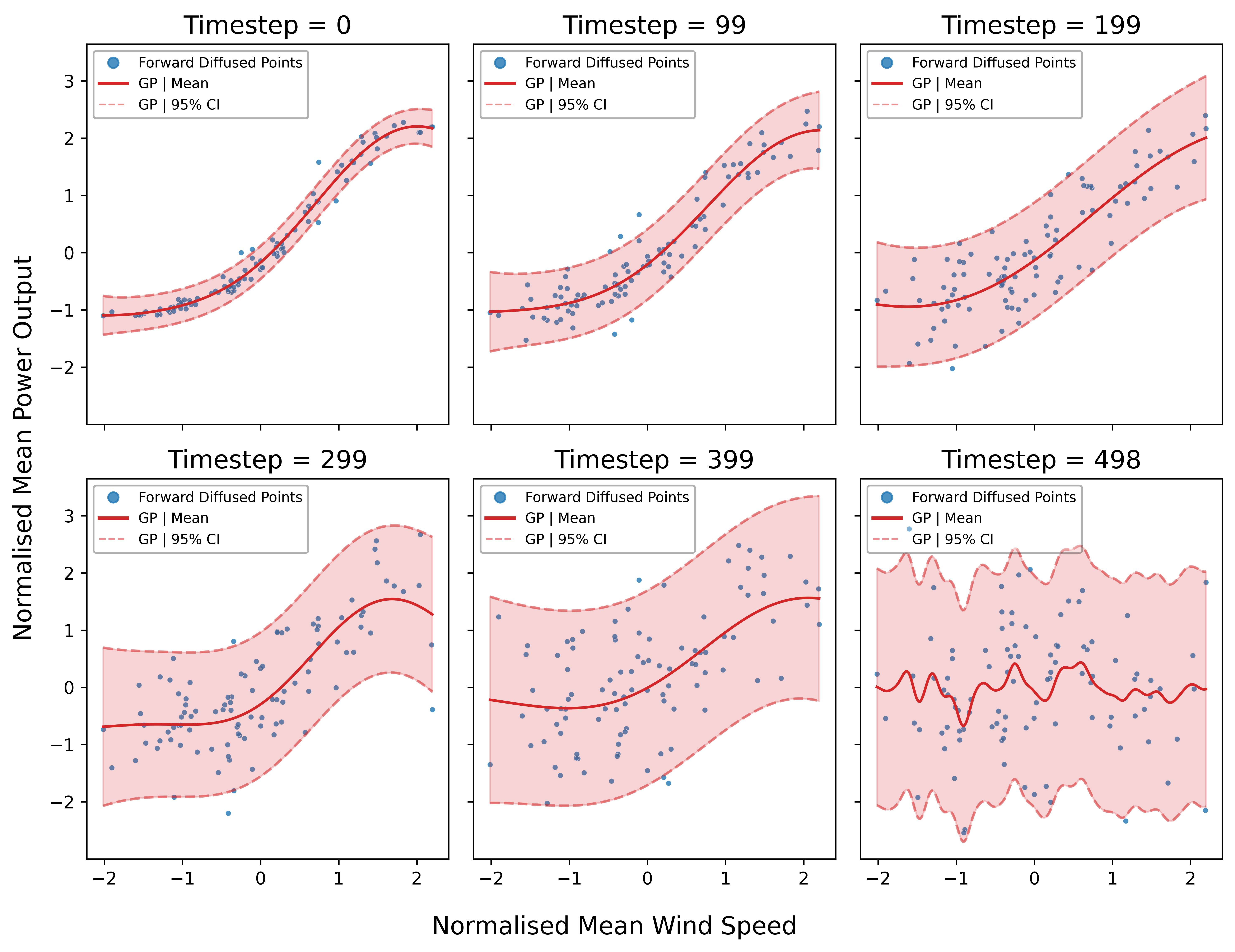}
\caption{Illustration of the forward diffusion process. Gaussian noise is added incrementally at each timestep under a cosine variance schedule, progressively degrading the structure of the data until it converges to white noise. GP: Gaussian process, CI: confidence interval.}
\label{fig:dd_f}
\end{figure}

We provide qualitative illustrations of the diffusion dynamics underlying NDPs, complementing the theoretical framework presented in Section~\ref{sec:ndp}. The aim is to build intuition for the forward corruption process and for the behaviour of unconditional and conditional sampling during reverse diffusion. We focus on a one-dimensional regression setting for clarity and visualise representative trajectories at different diffusion timesteps. For these illustrations, we use representative observations from a wind turbine power curve, with further details of the dataset provided in Section~\ref{subsec:datasets}. Fig.~\ref{fig:dd_f} illustrates the forward diffusion process, where Gaussian noise is gradually added to the outputs using a cosine variance schedule \cite{nichol2021improveddenoisingdiffusionprobabilistic}. As the timestep $t$ increases, the data loses structure and converges towards white noise. This process is applied independently at each input location while the inputs themselves remain fixed.

Unconditional sampling starts from random noise and follows the reverse diffusion process for a fixed number of timesteps. Even without any context information, the resulting samples already resemble the overall data distribution, indicating that the learned prior captures meaningful structure. For wind turbine data, where the relationships between inputs and power output are similar across turbines, the prior alone often produces plausible results.

\begin{figure}[!t]
\centering
\includegraphics[width=1\textwidth]{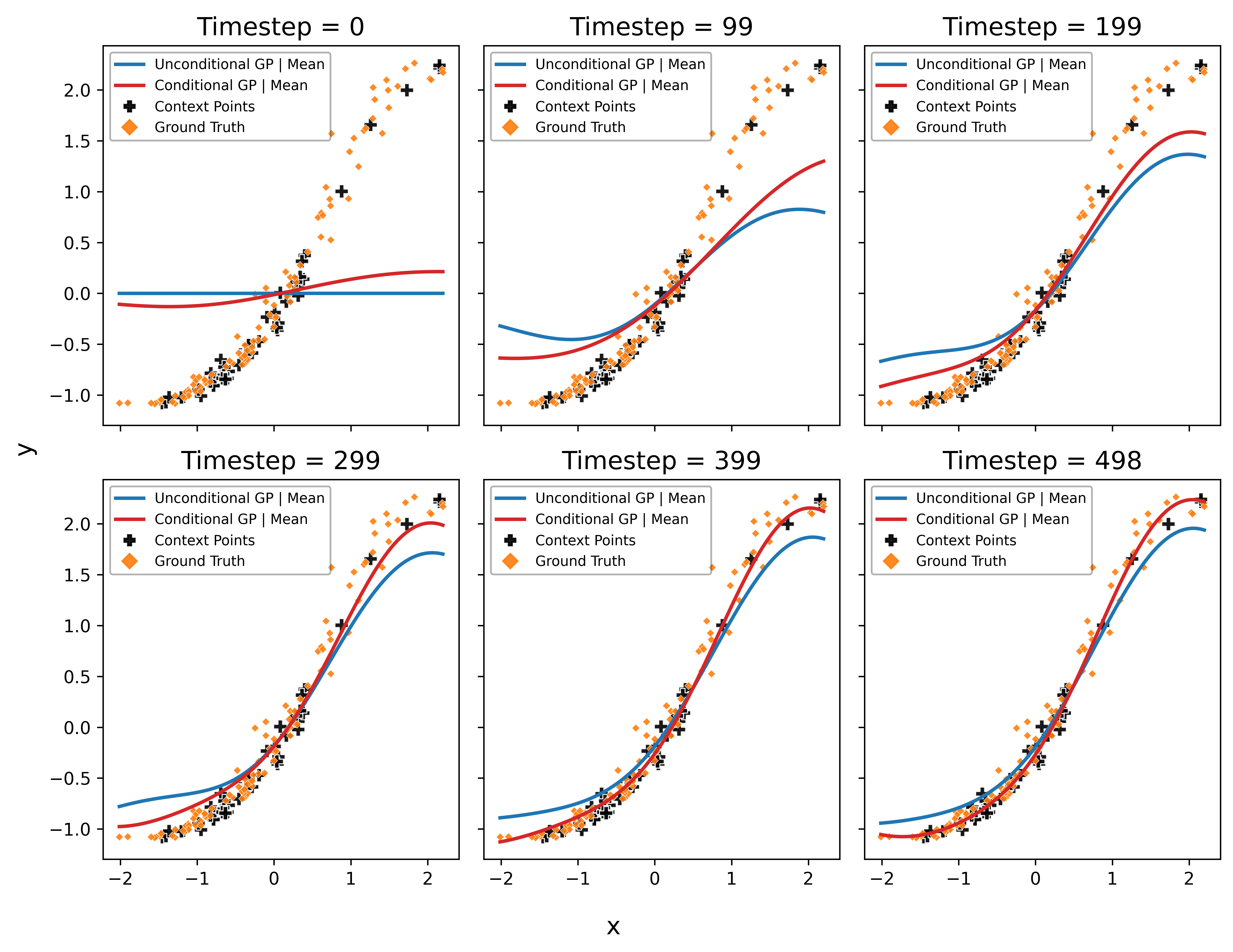}
\caption{Comparison of unconditional and conditional reverse diffusion sampling. We sampled 10 reverse trajectories per input location, plotting the mean of these. Conditional sampling provides a closer fit to the true data across the illustrated timesteps. GP: Gaussian process.}
\label{fig:dd_b_u_c}
\end{figure}

Conditional sampling incorporates context points at each timestep of the reverse process, guiding the trajectories towards functions that are consistent with the observed data. Initially, when $y^{*}_{t}$ is close to pure noise, the context points already contain useful structure. This pulls the reverse trajectories towards functions consistent with the context set, leading to samples that better fit the true data. Fig.~\ref{fig:dd_b_u_c} compares unconditional and conditional sampling. Even at early timesteps, the conditional trajectories are visibly influenced by the context points, producing outputs that more closely match the true data throughout the reverse process. 
Additional illustrations of the forward and reverse diffusion trajectories for an individual input point, under both conditional and unconditional sampling, are provided in Appendix~\ref{app:single_point_diffusion}.
\section{Multi-task Neural Diffusion Processes}
\label{sec:mtndp}
We propose a novel architectural addition to extend NDPs to MT-NDPs, building on ideas from \cite{kim2022multitaskneuralprocesses}. The broad idea is to add a task encoder that can take context points and use that information to identify the task the model is performing a prediction on. This information is represented in the encoding space by a single vector $v \in \mathbb{R}^\kappa$. That is, the encoding space has dimension $\kappa$, allowing enough flexibility for the model to distinguish between different tasks. The dimension $\kappa$ is assumed to be small; for example, in our experiments described in Section~\ref{sec:experimental_design}, we consider $\kappa \in \{4, 8, 16\}$. This vector can then be propagated downstream to the diffusion architecture, which remains unchanged. The way we pass the information downstream is simply to concatenate these extra $\kappa$ dimensions onto the other input features. These $\kappa$ additional features will be the same for all points in the same training (or testing) sample.

\subsection{Forward Diffusion  Process}

Let $x_0=x$ and $y_0=y$, and suppose we have
context observations $x^c=x_0^c$ and $y^c=y_0^c$.
Then MT-NDPs gradually add noise according to
\begin{equation}
    q\!\left(
    \begin{bmatrix}
    x_t \\
    y_t \\
    x^c_t \\
    y^c_t
    \end{bmatrix}
    \Bigg|
    \begin{bmatrix}
    x_{t-1} \\
    y_{t-1} \\
    x^c_{t-1} \\
    y^c_{t-1}
    \end{bmatrix}
    \right)
    = \mathcal{N}\!\left(
    y_t ; \sqrt{1 - \beta_t}\, y_{t-1}, \, \beta_t I
    \right)
    \label{eq:ndp_f1}
\end{equation} for each timestep $t$ where $\beta_t$ is the noise from the variance schedule. Notice the similarities to Eq.~\ref{eq:ndp_f}, where now, as well as explicitly differentiating between the input locations $x_t$ and outputs $y_t$, we also incorporate the context points $x^c_t$ and $y^c_t$. During training, Gaussian noise is added to the target outputs, while the input locations and the context observations used to construct the task representation remain unchanged. During RePaint-based conditional sampling, noisy versions of the context outputs are generated separately, as described in Section~\ref{subsec:mtndp_sampling}. By the final timestep $t = T$, the function values $y_t$ will look like Gaussian noise from $\mathcal{N}(0,1)$.

\subsection{Task Encoding and Reverse Diffusion}

MT-NDPs, like NDPs, use an NN that learns to denoise the corrupted function values, where we now also make use of some context points $(x_0^c,y_0^c)$.  The input locations $x_t$ and the original context observations $(x_0^c,y_0^c)$ used by the task encoder remain uncorrupted, while the target outputs $y_t$ are corrupted. In fact, the model will use the uncorrupted $x_t$ and
the original context observations $(x_0^c,y_0^c)$
to help improve the denoising model. Here, the task encoder uses the original context observations throughout the denoising process; noisy context observations are introduced separately during RePaint-based conditional sampling, as described in Section~\ref{subsec:mtndp_sampling}. The parametrised kernel is of the form:

\begin{equation}
    p_\theta\!\left(
    \begin{bmatrix}
    x_{t-1} \\
    y_{t-1} \\
    x^c_{t-1} \\
    y^c_{t-1} \\
    \end{bmatrix}
    \Bigg|
    \begin{bmatrix}
    x_t \\
    y_t \\ 
    x^c_{t} \\
    y^c_{t} \\
    \end{bmatrix}
    \right)
    = \mathcal{N}\!\left(
    y_{t-1} ; \mu_\theta(x_t, y_t, x^c_t, y^c_t, t), \, \tilde{\beta}_t I
    \right)
    \label{eq:mtndp_b}
\end{equation} with 
\begin{equation}
    \mu_\theta(x_t, y_t,  x^c_t, y^c_t, t)
    = \frac{1}{\sqrt{\alpha_t}}
    \left(
    y_t - \frac{\beta_t}{\sqrt{1 - \bar{\alpha}_t}} \,
    \epsilon^{MT}_\theta(x_t, y_t,  x^c_t, y^c_t, t)
    \right),
    \label{eq:ndp_mean}
\end{equation} where the noise model $ \epsilon^{MT}_\theta: \mathbb{R}^{( N + M ) \times D} \times \mathbb{R}^{(N + M)} \times \mathbb{R} \;\to\; \mathbb{R}^N$ , now leverages the input locations $x_t$ and context points $(x^c_t, y^c_t)$.  This model is in fact the composition of the encoder model $ \epsilon^{\text{encoder}}_\theta$ and the original NDP denoising model   $ \epsilon^{NDP}_\theta$,
\begin{equation}
    \epsilon^{MT}_\theta(x_t, y_t,  x^c_t, y^c_t, t) =  \epsilon^{NDP}_\theta \circ \epsilon^{\text{encoder}}_\theta (x_t, y_t,  x^c_t, y^c_t, t).\end{equation}
This encoder $\epsilon^{\text{encoder}}_\theta (x_t, y_t,  x^c_t, y^c_t, t)$ does the following: 
    \begin{itemize}
    \item[(i)] Acts as the identity on $(x_t, y_t)$, and maps the original context observations
$(x_0^c, y_0^c)$ to some $\kappa$-dimensional space: $ \mathbb{R}^{( N + M ) \times D} \times \mathbb{R}^{(N + M)} \times \mathbb{R} \;\to\; \mathbb{R}^{N \times D}  \times  \mathbb{R}^N  \times  \mathbb{R}^{\kappa\times M}$,
    \item[(ii)] Taking the mean over the image of all context points in this $\kappa$-dimensional space,  $\bar{v} \in \mathbb{R}^\kappa$ : $\mathbb{R}^{N \times D}  \times  \mathbb{R}^N  \times  \mathbb{R}^{\kappa\times M}\to\; \mathbb{R}^{N \times  D}  \times  \mathbb{R}^N \times  \mathbb{R}^\kappa$,
    \item[(iii)] Copying this mean vector, $\bar{v}$, $N$ times so that it can be concatenated with the other input features: $\mathbb{R}^{N \times  D}  \times  \mathbb{R}^N \times  \mathbb{R}^\kappa \;\to\; \mathbb{R}^{N \times  D}  \times  \mathbb{R}^N \times  \mathbb{R}^{\kappa \times N} =  \mathbb{R}^{N \times (D + \kappa)} \times \mathbb{R}^N$
\end{itemize}
Therefore, given the initial input locations and targets $T = (x_{0} \in \mathbb{R}^{N \times D}, y_{0} \in \mathbb{R}^{N})$ and context points  $C = (x^c_{0} \in \mathbb{R}^{M \times D}, y^c_{0} \in \mathbb{R}^{M})$, after the encoder we preserve $T$ and are left with $N$ copies of a vector $\bar{v} \in \mathbb{R}^\kappa$ that we concatenate to the other input features of $x_0$ to leave us with $D + \kappa$ dimensional input features, and a modified set of input locations and targets: $T^{*} = (x^{*}_{0} \in \mathbb{R}^{N \times ( D+ \kappa)}, y^{*}_{0} \in \mathbb{R}^{N})$ that is then passed through the NDP denoising model  $ \epsilon^{NDP}_\theta(x^{*}_{t}, y^{*}_{t}, t): \mathbb{R}^{N \times (D + \kappa)} \times \mathbb{R}^N \to\; \mathbb{R}^N$. By constructing the task encoder in this way, we ensure agnosticity to input size. Permutation invariance with respect to the ordering of context points is ensured by aggregating context information through the mean operator $\bar{v} \in \mathbb{R}^\kappa$. The target points are kept unchanged by the encoder. This construction is compatible with invariant noise model architectures, as discussed in Section~\ref{subsec:noise_model_arch}.

\subsection{Training Objective}

The objective is very similar to the NDP objective in Eq.~\ref{eq:ndp_loss}:
\begin{equation}
    L_\theta = 
    \mathbb{E}_{t, x_0, y_0, x^c_0, y^c_0,\epsilon}
    \left[
    \left\|
    \epsilon - \epsilon^{MT}_\theta(x_0, y_t, x^c_0, y^c_0, t)
    \right\|^2
    \right],
    \label{eq:ndp_loss1}
\end{equation} where  $x_0 = x_t$, $x^c_0 = x^c_t$, and $y^c_0 = y^c_t$ for all timesteps $t$ and  $y_t$ follows the same corruption process as in Eq.~\ref{eq:y_t}.

\subsection{Conditional and Unconditional Sampling} \label{subsec:mtndp_sampling}
Unconditional sampling remains unchanged from the NDP setup described in Section \ref{sec:ndp}, as we use no context points at all. Simply take $y_T  \sim \mathcal{N}(0, I)$, and then for $t = T, \dots, 1$ apply the parametrised kernel.  
For conditional sampling, we still follow the process for conditional sampling of NDPs described in Section \ref{sec:ndp}, but now using our upgraded denoising model $\epsilon^{MT}_\theta$: 

\begin{equation}
    y_{t-1} \sim \mathcal{N}\!\left( 
        \frac{1}{\sqrt{\alpha_{t}}}
        \left( y_{t} - \frac{\beta_{t}}{\sqrt{1 - \bar{\alpha}_{t}}}\, 
        \epsilon^{MT}_{\theta}(x_{0}, y_{t}, x^c_0, y^c_0, t) \right),
        \ \tilde{\beta}_{t} I
    \right).
    \label{eq:mtndp_cond}
\end{equation}  
During conditional sampling, the task encoder and RePaint use the
context observations in two distinct ways. The task encoder operates
on the original, uncorrupted context observations $(x_0^c,y_0^c)$
to construct the task-specific representation. In contrast, RePaint
generates noisy versions of the context outputs at each diffusion
timestep according to the forward diffusion process. These noisy
context outputs are concatenated with the current target values
before applying the reverse diffusion transition. RePaint also
performs repeated reverse and forward diffusion steps to improve
conditioning on the observed context. Thus, the original context
observations remain fixed for the task encoder, while their noisy
counterparts are used during RePaint-based conditional sampling.
\section{Experimental Design}
\label{sec:experimental_design}

We evaluate the proposed MT-NDP framework through three experimental studies designed to investigate its predictive performance, uncertainty quantification, and ability to generalise across related tasks. First, we consider a real-world wind energy application using SCADA data from multiple wind turbines. Second, we conduct controlled synthetic experiments to investigate the effect of the task representation dimensionality and the role of context size in generalisation to unseen tasks. Third, we evaluate MT-NDPs on a School Examination dataset, providing an additional real-world multi-task regression benchmark. Across the real-world data experiments, we assess predictive accuracy, uncertainty calibration, and performance relative to relevant neural-process and Gaussian-process baselines. 

\subsection{Datasets}
\label{subsec:datasets}
\noindent\textbf{Wind farm SCADA dataset.} Wind farm SCADA data provide a realistic and challenging benchmark for multi-task probabilistic regression. The data are high-frequency, noisy, and high-dimensional, and wind farms consist of multiple turbines that operate under broadly similar environmental conditions while exhibiting turbine-specific behaviour. This setting naturally motivates models that can share information across turbines while adapting to individual dynamics.  From a modelling perspective, this application places simultaneous demands on scalability, uncertainty calibration, and cross-task generalisation. Models must handle large datasets spanning multiple years, provide reliable predictive uncertainty, and adapt to turbines with limited historical data. In particular, the ability to condition predictions on a small number of turbine-specific observations at inference time is essential in realistic operational scenarios. 

We use SCADA data from the six Senvion MM92 wind turbines at Kelmarsh wind farm in the UK \cite{Plumley2022}. The dataset spans January 2016 to July 2021 at a ten-minute resolution, comprising more than 1.7 million records across 110 variables, including wind speed, wind direction, power output, and component temperatures. Each variable is reported as mean, minimum, maximum, and standard deviation within the ten-minute bins. Further details on SCADA variables and the complete data cleaning procedure are provided in the Appendix \ref{dataset}. All preprocessing steps are applied consistently across models.  We emphasise that the objective of this application is probabilistic wind power curve regression rather than
time-series forecasting. Specifically, we model the relationship between observed turbine operating conditions
and the corresponding power output, without explicitly
modelling temporal dependencies or predicting power
generation at future time horizons.

 \vspace{0.5em}
\noindent\textbf{Synthetic multi-task regression dataset.} We consider a synthetic multi-task regression dataset to investigate the ability of MT-NDPs to learn representations shared across related tasks and generalise to previously unseen tasks. The generated dataset comprises $25$ regression tasks, of which
$20$ are used for training and the remaining five are held out for evaluation. Each task is defined by
$f_j(x)=a_j\sin(\omega_jx+b_j)
+c_j|x-\tau_j|
+d_j\tanh(k_j(x-\tau_j)),$
where $a_j$, $\omega_j$, and $b_j$ control the amplitude,
frequency, and phase of the sinusoidal component,
respectively; $c_j$ controls the magnitude of the
non-smooth component; $\tau_j$ determines the change-point
location; and $d_j$ and $k_j$ control the magnitude and
sharpness of the smooth transition, respectively. Inputs are sampled uniformly from $[-3,3]$, with independent
Gaussian observation noise of standard deviation $0.1$. Both training and test evaluations are presented in Appendix Fig. \ref{fig:synthetic_example}. Each task corresponds to a distinct underlying function, allowing us to investigate how information learned from multiple related functions can be transferred to unseen tasks. This controlled setting provides a basis for examining the sensitivity of MT-NDPs to the dimensionality of the learned task representation and the number of available context observations.

  \vspace{0.5em}
\noindent\textbf{School examination dataset.} We use the Exam dataset \cite{goldstein1993multilevel}, which contains examination results from 4,059 students across 65 schools. The dataset includes normalised examination scores and student-level characteristics, such as prior attainment and gender, alongside school-level information. Its hierarchical structure makes it suitable for multi-task regression, where each school is treated as a separate task, allowing us to evaluate generalisation to previously unseen schools using limited context observations.

\subsection{Baseline Models}
\label{subsec:baselines}

For the real-world experiments, we compare MT-NDPs with pooled single-task neural diffusion process (ST-NDP) baselines trained on observations from the training tasks without an explicit task encoder. Unlike MT-NDPs, these baselines do not receive task-specific representations, although they can incorporate context observations through RePaint-based conditional sampling. All NDP variants share the same diffusion architecture, differing only in whether task-specific information is incorporated. Additional experiments investigating the effect of input dimensionality, using one-dimensional and five-dimensional feature sets, are reported in the Appendix \ref{res:1dvs5d}. We also investigate GP baselines as a classical probabilistic regression approach, with the corresponding experimental results presented in the Appendix \ref{res:gpvsndp}. While GPs provide principled uncertainty estimates in small-data regimes, they scale poorly with dataset size and dimensionality, and sparse approximations have been shown to yield underconfident uncertainty estimates on this dataset \cite{Fiocchi_2025}. GP models are trained using the same data splits and preprocessing as the NDP-based models. Theoretical background and further details for the GP baselines are provided in Appendix~\ref{subsec:gp}. We also consider multi-task neural processes (MT-NPs) \cite{kim2022multitaskneuralprocesses}, attentive neural processes (ANPs) \cite{kim2019attentiveneuralprocesses}, and multi-task Gaussian processes (MT-GPs) \cite{bonilla2007multi}.

\subsection{Experimental Setup and Implementation Details}
\label{subsec:experimental_protocols}
\label{subsec:modelparameters}
Some aspects of the experimental design are common to all comparisons and are therefore outlined here. NDPs learn distributions over functions, which requires training on many function samples. The baseline models were evaluated using the same
test observations and context sets as MT-NDP.
Further details on their architectures, training
configurations, and optimisation procedures are
provided in Appendix~\ref{app:baseline_implementation}. Inference times were measured for the complete prediction
procedure, including task conditioning and the generation
of predictive samples where applicable. Model training
and checkpoint loading were excluded from these measurements.

 \vspace{0.5em}
 \noindent\textbf{Wind energy experiments.}
In our setting, each function sample consists of $100$ covariate--target pairs. To reduce the impact of autocorrelation in wind generation, both training and testing samples are shuffled so that they are not sequential. For these samples to represent meaningful draws from the underlying stochastic process, all points in a given sample are taken from the same turbine. Since our objective is to model the wind power curve and capture dependencies across turbines rather than perform time-series forecasting, temporal ordering is not explicitly considered in the modelling framework.

To evaluate generalisation to previously unseen turbines, we consider two experimental configurations. For the main results, models are trained on turbines~1--5 and evaluated on turbine~6. For the supplementary experiments presented in Appendix~\ref{res:1dvs5d} and Appendix~\ref{res:gpvsndp}, models are trained on turbines~2--6 and evaluated on turbine~1. Context points, drawn from the training split of the corresponding test turbine, are provided during inference to guide predictions. These context observations are kept separate from the test functions, ensuring that evaluation targets are not used for conditioning. This setup allows us to assess (i) how informative the learned prior is in the absence of context points, and (ii) the benefit of incorporating context points for predictive accuracy and uncertainty calibration.

All features are standardised per wind turbine, using the mean and standard deviation from the wind turbine's training set. This step is necessary for all NDP variations and baseline models. To ensure fair comparisons, we use the same test functions across all models and context sizes. Context sets of size $0$, $25$, and $50$ points are considered. During training, the number of context points in each set is sampled uniformly from the integers $0$ to $50$. This setting exposes the model to situations with no context, with few context points, and with many context points. This allows the model to learn a shared distribution across tasks while adapting its predictions to individual turbines when context observations are available. The design reflects realistic deployment scenarios, where the amount of available historical data can vary substantially across turbines. Each evaluation uses $30$ test functions of length $100$. For MT-NDPs, predictions without context observations are generated through unconditional sampling, whereas predictions with context observations are obtained using RePaint-based conditional sampling, as described in Section~\ref{subsec:mtndp_sampling}. The same evaluation data and context observations are used across the competing baselines.

For the wind-energy NDP experiments, we use a cosine noise schedule \cite{nichol2021improveddenoisingdiffusionprobabilistic} with $T=500$ diffusion steps. The denoising NN has $4$ hidden layers of dimension $64$, with multi-head self-attention using $8$ attention heads. The optimiser is Adam \cite{Kingma2014ADAM:OPTIMIZATION}, following the configuration of \cite{dutordoir2023neural}, with a $20$-epoch warmup, $200$-epoch decay, learning rate schedule ($2 \times 10^{-5} \to 10^{-3} \to 10^{-5}$), and an EMA weight of $0.995$. For sampling, we generate $200$ reverse diffusion trajectories in the one-dimensional case and $100$ in higher-dimensional settings, balancing predictive accuracy and computational cost.

In the multi-task framework, the task encoder is a multilayer perceptron (MLP) with $4$ layers of size $64$, GeLU activations, and an $8$-dimensional output embedding ($\kappa=8$). The mean embedding across all context points forms a task representation vector. Training is performed for $250$ epochs using batches of $32$ function samples of length $100$, with $500$ samples drawn per wind turbine ($5$ tasks/turbines in total), and $100$ randomly shuffled function samples presented per epoch. Note that the MLP-based task encoder is not invariant to permutations of input features. However, this is less restrictive in the wind-energy application, where the ordering of input features remains fixed across training and testing.

\vspace{0.5em}
 \noindent\textbf{Synthetic ablation study.} Training samples contain $50$ observations, while each
held-out task is evaluated on $100$ test points. For each held-out task, context observations are selected separately from the $100$ evaluation points, ensuring that the observations used for conditioning are not included in the predictive accuracy or coverage calculations. We investigate embedding dimensions
$\kappa\in\{4,8,16\}$ and context sizes
$C\in\{0,5,10,25\}$.
Each configuration is evaluated over two repetitions,
using the same held-out tasks and evaluation points. The MT-NDP uses a cosine noise schedule with $T=50$
diffusion steps and is trained using Adam with a learning
rate of $10^{-4}$ and batch size $8$. Training is limited
to $10{,}000$ optimisation steps, with validation every
$100$ steps and early stopping after $10$ consecutive
validation checks without an improvement exceeding
$10^{-3}$. Predictions are generated using $100$
reverse diffusion trajectories, with RePaint-based
conditioning applied when context observations are
available. All the experiments were conducted using an NVIDIA RTX PRO 6000
Blackwell Server Edition GPU.

\vspace{0.5em}
 \noindent\textbf{School Examination experiments.} We retain schools with at least $50$ observations,
treating each school as a separate regression task.
This threshold ensures that sufficient observations
are available for both conditioning and evaluation,
as we use $20$ observations as context points. In the first experiment, we use $30$ schools for training,
$5$ for validation, and report results for $6$ held-out
schools. For each held-out school, we select $20$ context
observations and evaluate predictions on the remaining
observations. We compare MT-NDPs against ST-NDPs without the task
encoder, MT-NP, ANP, and MT-GP baselines. In a second experiment, we perform leave-one-school-out
evaluation across $6$ schools, training on $5$ schools
and testing on the remaining school in each iteration.
The results of this additional experiment are presented
in Appendix~\ref{app:schools}. The school experiments use $T=50$ diffusion steps,
with a denoising network comprising $4$ layers of
width $64$ and $4$ attention heads. The task embedding
dimension is set to $\kappa=16$ in the first experiment
and $\kappa=4$ in the leave-one-school-out experiment.
Conditional predictions are generated using $100$
reverse diffusion trajectories. The experiments were conducted using the same GPU
configuration as described in the synthetic ablation study.

\subsection{Evaluation Metrics}
\label{subsec:model_evaluation}

We evaluate both point accuracy and predictive uncertainty. The same evaluation metrics are used across the wind-energy, synthetic, and school examination experiments, allowing consistent assessment of predictive accuracy and uncertainty calibration.

For a given test sample $s$, let $\{(x_i,y_i)\}_{i=1}^{N_s}$ denote inputs and observed outputs, where $N_s$ is the number of evaluation observations in sample $s$; $\hat{y}_i$ is the model's point prediction. For NDP/MT-NDP models, point predictions are the empirical mean over reverse-diffusion trajectories at each $x_i$, and central prediction intervals are the corresponding empirical quantiles across trajectories. For the other probabilistic models, point predictions and predictive intervals are obtained from their corresponding predictive distributions.

To assess point prediction accuracy, we compute the following per-sample losses for each test set: (i) mean absolute error (MAE), defined as $\mathrm{MAE}_s = \tfrac{1}{N_s}\sum_{i=1}^{N_s}\lvert y_i-\hat{y}_i\rvert$, and (ii) root mean squared error (RMSE), defined as $\mathrm{RMSE}_s = \sqrt{\tfrac{1}{N_s}\sum_{i=1}^{N_s}(y_i-\hat{y}_i)^2}$.

To evaluate predictive uncertainty, we compute the coverage error (CE), which quantifies the discrepancy between the nominal coverage $q$ and the actual coverage achieved by the predictive intervals:
\begin{equation}
CE_s \;=\; \frac{1}{|Q|}\sum_{q \in Q}
\left|\, \frac{1}{N_s}\sum_{i=1}^{N_s}
\mathbf{1}\{\, y_i^{(s)} \in I_q^{(s)}(x_i)\,\} - q \,\right|,
\end{equation}
where $Q$ is the set of nominal coverage levels and $q \in Q$ denotes a particular level, $y_i^{(s)}$ and $N_s$ denote the observed outputs and number of points respectively in sample $s$, and the term $I_q^{(s)}(x_i)$ is the predicted interval at coverage level $q$ for that input. The indicator function $\mathbf{1}\{\cdot\}$ equals $1$ if the condition inside holds and $0$ otherwise. The inner average computes the empirical coverage at level $q$, subtracting $q$ gives the deviation from the nominal level, and the absolute value penalises both under- and over-coverage equally. Averaging across all $q \in Q$ yields the mean coverage error $CE_s$, with smaller values indicating better calibrated predictive intervals. Hence, a good model yields $CE_s \approx 0$, indicating predictive intervals that closely align with their nominal coverage.

For the wind-energy experiments, results are averaged over $30$ test function samples, each containing $100$ input--output pairs, for context sizes $C\in\{0,25,50\}$. The corresponding numerical summaries are reported in Tables~\ref{tbl:results_turbine6} and~\ref{tbl:results_turbine1}. For the synthetic ablation study, the same metrics are evaluated across the $5$ previously unseen tasks, for embedding dimensions $\kappa\in\{4,8,16\}$ and context sizes $C\in\{0,5,10,25\}$, with results summarised over two repetitions. For the school examination experiments, the metrics are computed on the held-out observations of each test school, excluding the $20$ context observations used for conditioning. Results are then summarised across the corresponding held-out schools.

\section{Experimental Results}
\label{sec:results}
\subsection{Wind Power Prediction} \label{sec:res_mtndp}
We assess the benefits of extending NDPs to the multi-task setting. We evaluate two variants: an ST-NDP with five input features and no task encoder, and our proposed MT-NDP, which incorporates a task encoder and RePaint-based conditional diffusion. To provide a broader comparison with existing probabilistic regression approaches, we additionally evaluate MT-NPs, ANPs and MT-GPs. These benchmarks allow us to assess the predictive accuracy and uncertainty calibration of MT-NDPs relative to established neural process and Gaussian process approaches. We also conduct two supplementary experiments to further investigate the behaviour of NDPs on wind energy data. First, we compare a one-dimensional NDP with a GP baseline using wind speed as the sole input feature, focusing on predictions without context observations. Second, we investigate the effect of input dimensionality by comparing one-dimensional and five-dimensional NDPs. The experimental settings and corresponding results are presented in Appendix~\ref{sec:additional_wind}.

\begin{figure}[!ht]\begin{center}\includegraphics[width=1\textwidth]{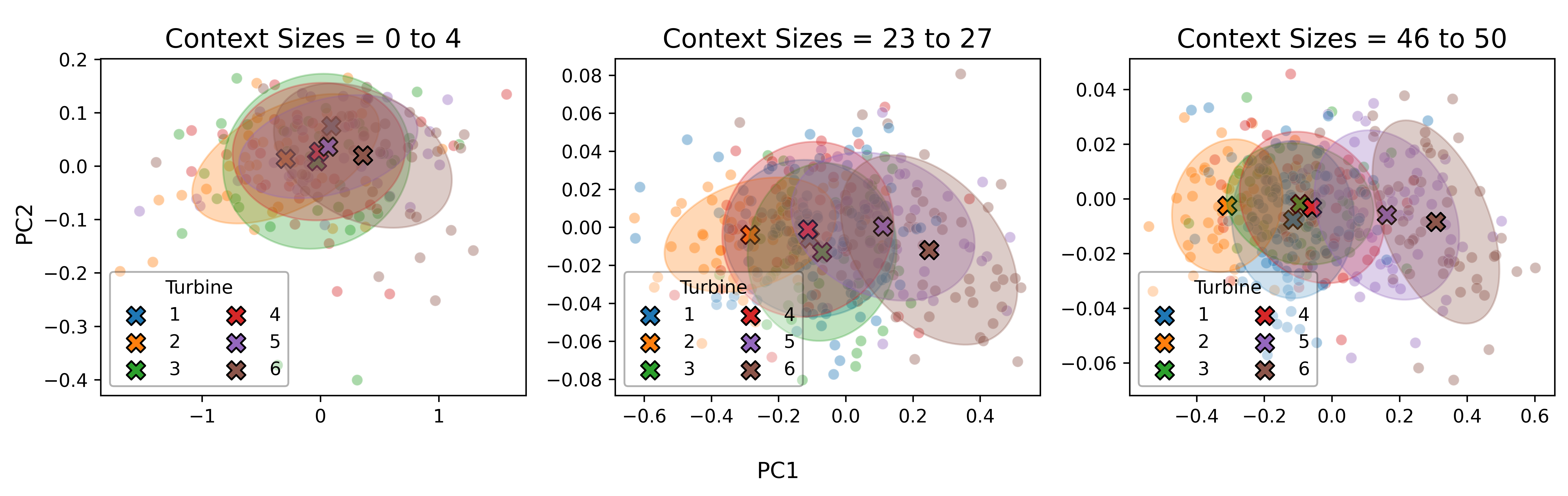}
        \caption{Principal component (PC) projection of encoder embeddings for turbines $1$–$6$, trained on turbines $2$–$6$. Each panel corresponds to a different range of context set sizes. Colours denote turbines, shaded ellipses show $68$\% coverage regions, and crosses mark cluster means. As context increases, the encoder begins to separate turbines, with turbine~$6$ emerging as the most distinct.
        \label{fig:enc_1}}
    \end{center}
\end{figure}

 \vspace{0.5em}
 \noindent\textbf{Encoder analysis.}
To understand how the encoder represents turbines, we project its $\kappa$-dimensional embeddings into two principal components using principal component analysis (PCA). Fig.~\ref{fig:enc_1} shows the resulting clusters when the model is trained on turbines $2$–$6$ and tested on turbine~$1$, with context set sizes ranging from very small ($0$–$4$ points) to large ($46$–$50$ points). With few context points (left panel), wind turbines are essentially indistinguishable, reflecting the fact that the model has little task-specific information and must rely on its global prior. As context increases (middle and right panels), the encoder begins to separate turbines, with turbine $6$ in particular forming a clearly distinct cluster. By contrast, turbine $1$ remains centred near the origin across all context sizes.   
 These findings suggest that the learned task representations capture differences in turbine-specific behaviour as context information becomes available. In particular, turbine 6 exhibits a more distinctive representation, whereas turbine 1 remains closer to the centre of the embedding space. The main quantitative evaluation considers turbine 6 as a held-out task, providing an opportunity to assess adaptation to turbine-specific characteristics using limited context observations. For the evaluation on turbine 6, the models were retrained using turbines 1–5. Complementary experiments with turbine 1 as the held-out task are reported in Appendix~\ref{res:1dvs5d} and Appendix~\ref{res:gpvsndp}, allowing us to examine generalisation under a different train--test configuration.

\begin{figure}[!t]
    \centering
    \includegraphics[width=1\textwidth]{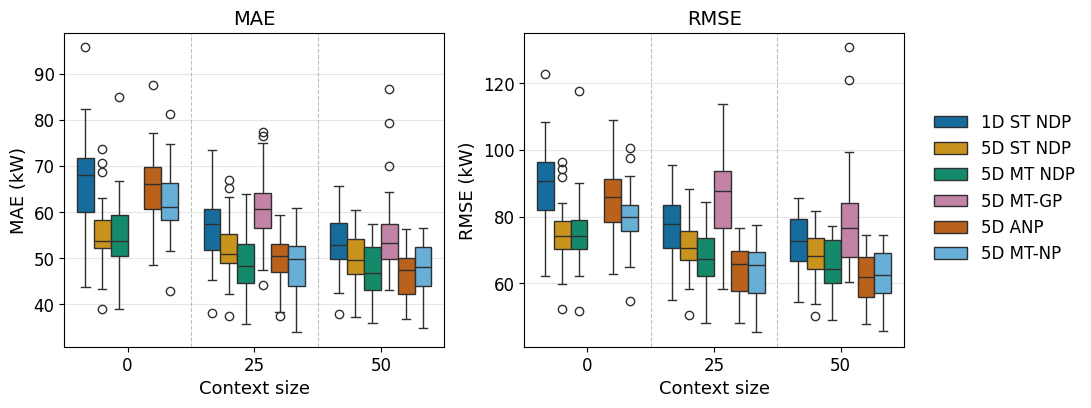}
    \caption{Mean absolute error (MAE, left) and root mean squared error (RMSE, right) distributions for the one-dimensional single-task neural diffusion process (1D ST-NDP), five-dimensional single-task neural diffusion process (5D ST-NDP), five-dimensional multi-task neural diffusion process (5D MT-NDP), multi-task Gaussian process (5D MT-GP), attentive neural process (5D ANP), and multi-task neural process (5D MT-NP) on turbine~6 across context sizes $0$, $25$, and $50$, aggregated over $30$ test functions of $100$ observations each. MT-GP results are not reported for context size equal to $0$, as the implementation requires task-specific observations to condition predictions for an unseen task.}
    \label{fig:paper_loss_t6}
\end{figure}

 \vspace{0.5em}
\noindent\textbf{Model comparison.} We compare the predictive performance of the one-dimensional ST-NDP, five-dimensional ST-NDP, and five-dimensional MT-NDP with three additional probabilistic regression approaches: MT-NPs, ANPs, and MT-GPs. Note that results for MT-GP are not reported for context size equal to $0$, as the implementation requires task-specific observations to condition predictions for an unseen task. Fig.~\ref{fig:paper_loss_t6} presents the distributions of MAE and RMSE on turbine~6 across different context sizes, evaluated over $30$ test functions of $100$ observations each. Prediction errors generally decrease as more context observations become available. Among the NDP variants, the five-dimensional MT-NDP achieves the lowest average MAE and RMSE at context sizes $25$ and $50$, highlighting the benefits of incorporating task-specific representations relative to single-task diffusion models. However, the additional NP benchmarks achieve competitive predictive accuracy, with MT-NP obtaining the lowest average MAE and RMSE at context size $25$ and ANP achieving the lowest values at context size $50$. The overlapping distributions also indicate variability in predictive performance across test functions.

\begin{figure}[!t]
    \centering
    \includegraphics[width=0.85\textwidth]{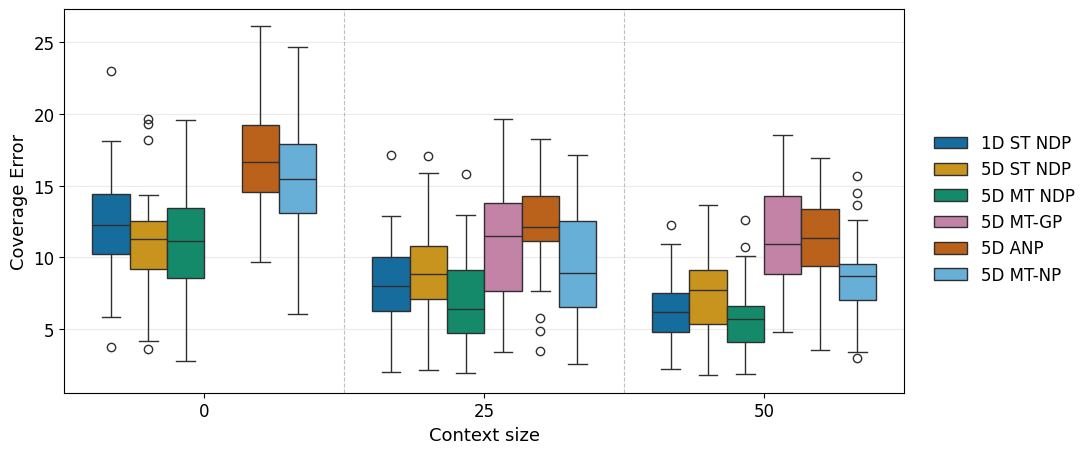}
    \caption{Coverage error (CE) distributions for the one-dimensional single-task neural diffusion process (1D ST-NDP), five-dimensional single-task neural diffusion process (5D ST-NDP), five-dimensional multi-task neural diffusion process (5D MT-NDP), multi-task Gaussian process (5D MT-GP), attentive neural process (5D ANP), and multi-task neural process (5D MT-NP) on turbine~6 across context sizes $0$, $25$, and $50$, aggregated over $30$ test functions of $100$ observations each. MT-GP results are not reported for context size equal to $0$, as the implementation requires task-specific observations to condition predictions for an unseen task.}
    \label{fig:paper_cov_err_t6}
\end{figure}

We next examine predictive uncertainty calibration. Fig.~\ref{fig:paper_cov_err_t6} presents the coverage error distributions across models and context sizes. The five-dimensional MT-NDP demonstrates favourable uncertainty calibration, particularly when context observations are available. At context sizes $25$ and $50$, MT-NDP achieves the lowest average CE among all evaluated models, with values of $6.986\%$ and $5.691\%$, respectively. Although MT-NP and ANP achieve slightly better point prediction accuracy in some settings, their higher coverage errors highlight the trade-off between predictive accuracy and uncertainty calibration. The corresponding empirical versus nominal coverage curves are provided in Appendix~\ref{sec:wind_coverage}.

Table~\ref{tbl:results_turbine6} provides the corresponding numerical summaries. At context size $50$, ANP achieves the lowest MAE ($46.860$ kW) and RMSE ($61.800$ kW), whereas MT-NDP achieves the lowest CE ($5.691\%$). Overall, the results indicate that MT-NDP achieves competitive predictive accuracy while providing the lowest average coverage error among the evaluated models when context observations are available. This suggests that the multi-task diffusion architecture is particularly effective at producing calibrated predictive distributions, even when alternative neural process approaches achieve slightly lower point prediction errors.

\begin{table}[!t]
\centering
\small
\setlength{\tabcolsep}{3pt}
\caption{Out-of-sample performance of single-task neural diffusion process (ST-NDP), multi-task neural process (MT-NP), multi-task Gaussian process (MT-GP), attentive neural process (ANP) and multi-task neural diffusion process (MT-NDP) models on turbine~$6$, averaged over $30$ test functions ($100$ points each). 
Metrics reported: mean absolute error (MAE, kW), root mean squared error (RMSE, kW), and coverage error (CE, \%). CE denotes the mean absolute coverage error across the evaluated nominal coverage levels, expressed in percentage points.
Bold values indicate the best overall performance across all models and context sizes. MT-GP results are not reported for context size equal to $0$, as the implementation requires task-specific observations to condition predictions for an unseen task. 1D: one-dimensional, 5D: five-dimensional.}
\label{tbl:results_turbine6}
\begin{tabular}{lccccccccc}
\toprule
& \multicolumn{3}{c}{Context = $0$} 
& \multicolumn{3}{c}{Context = $25$} 
& \multicolumn{3}{c}{Context = $50$} \\
\cmidrule(lr){2-4} \cmidrule(lr){5-7} \cmidrule(lr){8-10}
Model 
& MAE & RMSE & CE 
& MAE & RMSE & CE 
& MAE & RMSE & CE \\
\midrule \midrule
1D ST-NDP 
& $66.539$ & $90.491$ & $13.052$ 
& $56.641$ & $77.766$ & $8.779$ 
& $53.283$ & $73.429$ & $6.865$ \\
5D ST-NDP 
& $55.169$ & $75.441$ & $11.225$ 
& $51.864$ & $71.306$ & $9.058$ 
& $49.676$ & $68.465$ & $7.376$ \\

5D MT-GP

& \text{--} & \text{--} & \text{--}

& $61.120$ & $86.372$ & $11.184$

& $55.906$ & $79.639$ & $11.480$ \\

5D MT-NP
& $62.017$ & $79.826$ & $15.536$
& $48.974$ & $64.387$ & $9.332$
& $47.942$ & $63.269$ & $8.717$ \\

5D ANP
& $66.020$ & $85.680$ & $16.780$
& $49.300$ & $64.610$ & $11.950$
& $\mathbf{46.860}$ & $\mathbf{61.800}$ & $11.230$ \\

5D MT-NDP 
& $55.291$ & $75.974$ & $11.375$ 
& $49.183$ & $67.805$ & $6.986$ 
& $47.268$ & $65.636$ & $\mathbf{5.691}$ \\
\bottomrule
\end{tabular}
\end{table}

Regarding computational efficiency, we additionally examined the runtime required to complete approximately $100$ evaluation batches on turbine~6. The estimated runtime for the five-dimensional MT-NDP with RePaint was $114.20$ hours (approximately $4.76$ days),
compared with $60.25$ hours ($2.51$ days) for the five-dimensional ST-NDP with RePaint and $12.38$ hours ($0.52$ days) for the five-dimensional MT-NDP without RePaint. Thus, MT-NDP with RePaint required approximately $9.2$ times longer than its unconditional counterpart and $1.9$ times longer than ST-NDP with RePaint. These results highlight the computational cost associated with RePaint-based conditional sampling, particularly in the multi-task setting, and illustrate the trade-off between predictive uncertainty calibration and computational efficiency.

\subsection{Ablation Study}

\begin{figure}[!ht]
\begin{center}
\includegraphics[width=1\textwidth]{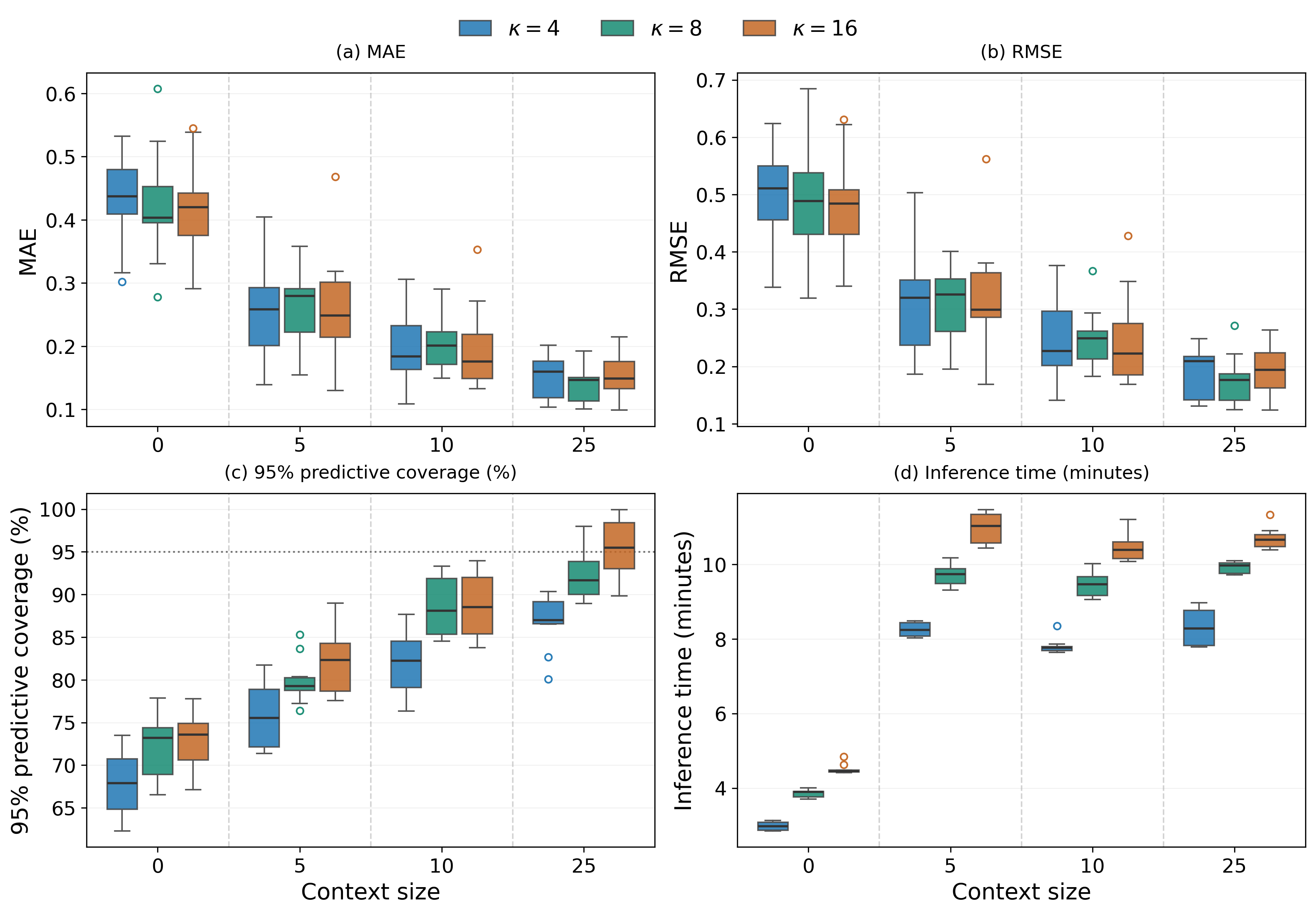}
\caption{Synthetic ablation study across embedding dimensions $\kappa$ and context sizes $C$. Panels show (a) mean absolute error (MAE), (b) root mean squared error (RMSE), (c) empirical coverage of nominal 95\% predictive intervals, and (d) inference time in minutes. Each boxplot represents ten evaluations. The dotted line in panel (c) indicates nominal 95\% coverage.}
\label{fig:synthetic_ablation}
\end{center}
\end{figure}

Using our synthetic dataset, we investigate the effects of task embedding dimension $\kappa \in \{4,8,16\}$ and context size $C \in \{0,5,10,25\}$ on predictive accuracy, uncertainty calibration, and computational efficiency. Fig.~\ref{fig:synthetic_ablation} presents the results across five unseen tasks and two repetitions, with numerical summaries provided in Appendix Table~\ref{tab:synthetic_ablation}. Fig.~\ref{fig:synthetic_ablation} shows that increasing context size generally improves predictive accuracy across all embedding dimensions, with the lowest MAE and RMSE observed at $C=25$. At this context size, $\kappa=8$ achieves the lowest MAE ($0.140$) and RMSE ($0.176$), compared with MAE values of $0.151$ for both $\kappa=4$ and $\kappa=16$, suggesting that increasing the embedding dimension beyond $8$ does not provide further improvements in point prediction accuracy in this setting. Empirical coverage also increases with context size. At $C=25$, coverage increases from $86.7\%$ for $\kappa=4$ to $92.2\%$ for $\kappa=8$ and $95.6\%$ for $\kappa=16$, with the latter being closest to the nominal $95\%$ level. However, larger embeddings increase inference time, with corresponding mean times of $8.31$, $9.91$, and $10.69$ minutes. Overall, these results suggest that context size has a stronger influence on predictive accuracy than embedding dimensionality in this experiment, while larger embeddings can improve interval coverage at the expense of computational efficiency.

\subsection{Generalisation to Unseen Schools}
\label{sec:schools_results}

To further evaluate the generalisation capabilities of MT-NDP
beyond the wind energy application, we consider the School
Examination dataset. In the first experiment, 30 schools are
used for training, five for validation, and six previously
unseen schools for testing. Each test school is evaluated
using $C=20$ context observations. Table~\ref{tab:schools_results}
compares MT-NDP against ST-NDP, MT-NP, ANP, and MT-GP. MT-NDP achieves an MAE of $0.564$ and an RMSE of $0.715$,
outperforming ST-NDP, which obtains an MAE of $0.634$
and an RMSE of $0.790$. Its predictive accuracy is also
competitive with the remaining baselines, although MT-NP
achieves the lowest MAE and RMSE.
In terms of uncertainty quantification, MT-NDP achieves
empirical coverage of $95.31\%$, closest to the nominal
$95\%$ level. However, ANP achieves the lowest calibration
error ($4.022$), compared with $4.940$ for MT-NDP for the full distribution.
The main computational limitation of MT-NDP is its
inference time, averaging $525.99$ seconds per test school,
compared with $333.84$ seconds for ST-NDP and less
than two seconds for the remaining baselines. An additional leave-one-school-out experiment involving
six schools is presented in Appendix~\ref{app:schools}.
Despite the smaller number of training schools, MT-NDP
maintains empirical coverage close to the nominal
$95\%$ level across all the leave-one-out schools.

\begin{table}[htbp]
\centering
\caption{Predictive performance on the School Examination dataset using 30 schools for training, five for validation, and six previously unseen schools for testing. Each test school is evaluated using $C=20$ context observations. Results are reported as mean (standard deviation) across the six test schools. MAE denotes mean absolute error, RMSE denotes root mean squared error, Coverage 95\% refers to the empirical coverage of nominal 95\% predictive intervals, and CE denotes calibration error. Time (s) represents the inference time in seconds per test school. Lower values of MAE, RMSE, CE, and inference time indicate better performance, while empirical coverage closer to the nominal 95\% level indicates better interval calibration. Bold values indicate the best performance for each metric.}
\label{tab:schools_results}
\small
\setlength{\tabcolsep}{4pt}
\begin{tabular}{lccccc}
\toprule
Model
& MAE 
& RMSE 
& Coverage 95\% 
&  CE (\%)
& Time (s)  \\
\midrule 
MT-NDP 
& 0.564 (0.069)
& 0.715 (0.080)
& \textbf{95.31 (3.22)}
& 4.940 (2.616)
& 525.99 (19.23) \\

ST-NDP 
& 0.634 (0.120)
& 0.790 (0.133)
& 91.76 (4.00)
& 6.377 (3.657)
& 333.84 (52.31) \\

MT-NP
& \textbf{0.562 (0.067)}
& \textbf{0.707 (0.071)}
& 92.64 (4.01)
& 4.161 (1.397)
& \textbf{0.17 (0.19)} \\

ANP
& 0.579 (0.058)
& 0.721 (0.071)
& 93.81 (3.20)
& \textbf{4.022 (1.865)}
& 0.80 (1.69) \\

MT-GP
& 0.599 (0.047)
& 0.742 (0.052)
& 93.26 (4.08)
& 4.032 (2.094)
& 1.76 (1.31) \\

\bottomrule
\end{tabular}
\end{table}
\section{Discussion}
\label{sec:discussion}
We examine the role of task conditioning in diffusion-based function models and its effect on generalisation and uncertainty estimation in multi-task regression. Our evaluation across wind turbine SCADA data, synthetic multi-task regression problems, and school examination outcomes demonstrates the applicability of MT-NDPs across different regression settings. In both these experiments, task conditioning enables the model to exploit information from related tasks while adapting to previously unseen tasks using limited context observations. In both real-world applications, MT-NDP improves predictive accuracy over its single-task counterpart and is competitive with established multi-task neural-process and Gaussian-process baselines.Introducing a task encoder allows the diffusion model to learn task-specific representations from context observations while leaving the underlying diffusion process unchanged. This separation between diffusion-based function modelling and task encoding allows task-specific information to be incorporated without modifying the underlying generative process. The synthetic ablation study further demonstrates that increasing context size generally produces greater improvements in predictive accuracy than increasing the dimensionality of the task embedding. Although larger embeddings can improve predictive interval coverage, these improvements are not necessarily accompanied by reductions in prediction error. These findings highlight the importance of context availability and the trade-off between predictive accuracy, uncertainty calibration, and computational complexity when selecting the task representation dimension.

Several limitations point to directions for future research. The task encoder is deliberately simple and relies on an MLP followed by mean pooling to aggregate context observations into a fixed-dimensional representation. Although mean pooling ensures permutation invariance with respect to context ordering, the encoder does not enforce invariance to permutations of input features. This is less restrictive in the applications considered, where input features have fixed semantic interpretations. The embedding-dimensionality ablation also suggests that simply increasing the representation size does not consistently improve predictive performance. Future work could therefore investigate more expressive task encoders, including attention-based or hierarchical architectures, to capture richer relationships between context observations without relying solely on higher-dimensional embeddings. A second limitation concerns computational efficiency. MT-NDP requires iterative reverse-diffusion sampling, resulting in substantially higher inference costs than neural-process and Gaussian-process baselines. This cost is particularly pronounced with RePaint, which introduces additional forward and reverse diffusion steps. Although diffusion-based sampling provides flexibility in modelling predictive distributions, its computational requirements may limit applications requiring rapid predictions. Future work could investigate accelerated diffusion samplers, reduced-step denoising procedures, and more efficient conditioning mechanisms to reduce inference time while preserving predictive flexibility. Finally, although our evaluation includes synthetic experiments and two distinct real-world applications, further investigation on larger and more heterogeneous task collections would help establish the generalisability of MT-NDPs across different levels of task similarity.

\section{Conclusion}
\label{sec:conclusion}

We introduced MT-NDPs as an extension of diffusion-based function models for task-conditioned probabilistic regression. By augmenting NDPs with a lightweight task encoder, the proposed framework supports few-shot adaptation across related tasks while providing flexible predictive uncertainty estimates. We evaluated MT-NDPs through synthetic experiments and two real-world applications involving wind turbine SCADA data and school examination outcomes. Across these settings, the results demonstrate the benefits of task conditioning for predictive accuracy and uncertainty quantification, particularly as the number of available context observations increases. The synthetic experiments highlight the influence of context size and task representation dimensionality, while the real-world applications demonstrate competitive predictive performance against established probabilistic regression methods and improvements over single-task diffusion baselines. Overall, our findings illustrate how task-specific information can be incorporated into diffusion-based regression without altering the underlying diffusion mechanism, establishing MT-NDPs as a flexible framework for multi-task probabilistic learning.

\bibliography{references} 

@book{Rasmussen2006GaussianLearning,
  title={Gaussian processes for machine learning},
  author={Williams, Christopher KI and Rasmussen, Carl Edward},
  year={2006},
  publisher={MIT press},
  address = {Cambridge, MA}
}

@article{goldstein1993multilevel,
  title={A multilevel analysis of school examination results},
  author={Goldstein, Harvey and Rasbash, Jon and Yang, Min and Woodhouse, Geoffrey and Pan, Huiqi and Nuttall, Desmond and Thomas, Sally},
  journal={Oxford review of education},
  volume={19},
  number={4},
  pages={425--433},
  year={1993},
  publisher={Taylor \& Francis}
}

@article{wang2016locally,
  title={Locally downscaled and spatially customizable climate data for historical and future periods for North America},
  author={Wang, Tongli and Hamann, Andreas and Spittlehouse, Dave and Carroll, Carlos},
  journal={PloS one},
  volume={11},
  number={6},
  pages={e0156720},
  year={2016},
  publisher={Public Library of Science San Francisco, CA USA}
}

@inproceedings{requeima2019gaussian,
  title={The gaussian process autoregressive regression model (gpar)},
  author={Requeima, James and Tebbutt, William and Bruinsma, Wessel and Turner, Richard E},
  booktitle={The 22nd International Conference on Artificial Intelligence and Statistics},
  pages={1860--1869},
  year={2019},
  organization={PMLR}
}

@article{bonilla2007multi,
  title={Multi-task Gaussian process prediction},
  author={Bonilla, Edwin V and Chai, Kian and Williams, Christopher},
  journal={Advances in neural information processing systems},
  volume={20},
  year={2007}
}

@article{alvarez2012kernels,
  title={Kernels for vector-valued functions: A review},
  author={Alvarez, Mauricio A and Rosasco, Lorenzo and Lawrence, Neil D},
  journal={Foundations and Trends{\textregistered} in Machine Learning},
  volume={4},
  number={3},
  pages={195--266},
  year={2012},
  publisher={Emerald Publishing Limited}
}

@inproceedings{dutordoir2023neural,
  title={Neural diffusion processes},
  author={Dutordoir, Vincent and Saul, Alan and Ghahramani, Zoubin and Simpson, Fergus},
  booktitle={International Conference on Machine Learning},
  pages={8990--9012},
  year={2023},
  organization={PMLR}
}

@article{caruana1997multitask,
  title={Multitask learning},
  author={Caruana, Rich},
  journal={Machine learning},
  volume={28},
  number={1},
  pages={41--75},
  year={1997},
  publisher={Springer}
}

@inproceedings{garnelo2018neuralprocesses,
  title={Conditional neural processes},
  author={Garnelo, Marta and Rosenbaum, Dan and Maddison, Christopher and Ramalho, Tiago and Saxton, David and Shanahan, Murray and Teh, Yee Whye and Rezende, Danilo and Eslami, SM Ali},
  booktitle={International conference on machine learning},
  pages={1704--1713},
  year={2018},
  organization={PMLR}
}

@inproceedings{kim2019attentiveneuralprocesses,
  title={Attentive Neural Processes},
  author={Kim, Hyunjik and Mnih, Andriy and Schwarz, Jonathan and Garnelo, Marta and Eslami, Ali and Rosenbaum, Dan and Vinyals, Oriol and Teh, Yee Whye},
  booktitle={International Conference on Learning Representations},
  year={2019},
  url={https://openreview.net}
}

@article{kim2022multitaskneuralprocesses,
  title={Multi-task neural processes},
  author={Kim, Donggyun and Cho, Seongwoong and Lee, Wonkwang and Hong, Seunghoon},
  journal={arXiv preprint arXiv:2110.14953},
  year={2021}
}

@inproceedings{sohldickstein2015deepunsupervisedlearningusing,
  title={Deep unsupervised learning using nonequilibrium thermodynamics},
  author={Sohl-Dickstein, Jascha and Weiss, Eric and Maheswaranathan, Niru and Ganguli, Surya},
  booktitle={International conference on machine learning},
  pages={2256--2265},
  year={2015},
  organization={PMLR}
}

@article{ho2020denoisingdiffusionprobabilisticmodels,
  title={Denoising diffusion probabilistic models},
  author={Ho, Jonathan and Jain, Ajay and Abbeel, Pieter},
  journal={Advances in neural information processing systems},
  volume={33},
  pages={6840--6851},
  year={2020}
}

@inproceedings{lugmayr2022repaintinpaintingusingdenoising,
  title={Repaint: Inpainting using denoising diffusion probabilistic models},
  author={Lugmayr, Andreas and Danelljan, Martin and Romero, Andres and Yu, Fisher and Timofte, Radu and Van Gool, Luc},
  booktitle={Proceedings of the IEEE/CVF conference on computer vision and pattern recognition},
  pages={11461--11471},
  year={2022}
}

@article{Kingma2014ADAM:OPTIMIZATION,
  title={Adam: A method for stochastic optimization},
  author={Kingma, Diederik P and Ba, Jimmy},
  journal={arXiv preprint arXiv:1412.6980},
  year={2014}
}

@article{Hadjoudj2023AOpportunities,
  title={A review on data-centric decision tools for offshore wind operation and maintenance activities: Challenges and opportunities},
  author={Hadjoudj, Yannis and Pandit, Ravi},
  journal={Energy Science \& Engineering},
  volume={11},
  number={4},
  pages={1501--1515},
  year={2023},
  publisher={Wiley Online Library}
}

@article{en13123132,
  title={Using SCADA data for wind turbine condition monitoring: A systematic literature review},
  author={Maldonado-Correa, Jorge and Mart{\'\i}n-Mart{\'\i}nez, Sergio and Artigao, Estefan{\'\i}a and G{\'o}mez-L{\'a}zaro, Emilio},
  journal={Energies},
  volume={13},
  number={12},
  pages={3132},
  year={2020},
  publisher={MDPI}
}

@book{kolmogorov1950,
  title={Foundations of the theory of probability: Second English Edition},
  author={Kolmogorov, Andre{\u\i} Nikolaevich},
  year={2018},
  publisher={Courier Dover Publications},
address ={NY, USA}
}

@article{Plumley2022,
  title={Kelmarsh wind farm data (0.0. 3)},
  author={Plumley, Charlie},
  journal={Zenodo, February},
  year={2022}
}

@inproceedings{nichol2021improveddenoisingdiffusionprobabilistic,
  title={Improved denoising diffusion probabilistic models},
  author={Nichol, Alexander Quinn and Dhariwal, Prafulla},
  booktitle={International conference on machine learning},
  pages={8162--8171},
  year={2021},
  organization={PMLR}
}

@article{yu2021,
  title={A regional wind power probabilistic forecast method based on deep quantile regression},
  author={Yu, Yixiao and Yang, Ming and Han, Xueshan and Zhang, Yumin and Ye, Pingfeng},
  journal={IEEE Transactions on Industry Applications},
  volume={57},
  number={5},
  pages={4420--4427},
  year={2021},
  publisher={IEEE}
}

@inproceedings{yu2019probabilistic,
  title={Probabilistic prediction of regional wind power based on spatiotemporal quantile regression},
  author={Yu, Yixiao and Han, Xueshan and Yang, Ming and Yang, Jiajun},
  booktitle={2019 IEEE industry applications society annual meeting},
  pages={1--16},
  year={2019},
  organization={IEEE}
}

@article{zhou2022,
  title={Performance improvement of very short-term prediction intervals for regional wind power based on composite conditional nonlinear quantile regression},
  author={Zhou, Yan and Sun, Yonghui and Wang, Sen and Mahfoud, Rabea Jamil and Alhelou, Hassan Haes and Hatziargyriou, Nikos and Siano, Pierluigi},
  journal={Journal of Modern Power Systems and Clean Energy},
  volume={10},
  number={1},
  pages={60--70},
  year={2021},
  publisher={SGEPRI}
}

@article{zhang2020,
  title={Improved deep mixture density network for regional wind power probabilistic forecasting},
  author={Zhang, Hao and Liu, Yongqian and Yan, Jie and Han, Shuang and Li, Li and Long, Quan},
  journal={IEEE Transactions on Power Systems},
  volume={35},
  number={4},
  pages={2549--2560},
  year={2020},
  publisher={IEEE}
}

@article{chen2014,
  title={Wind power forecasts using Gaussian processes and numerical weather prediction},
  author={Chen, Niya and Qian, Zheng and Nabney, Ian T and Meng, Xiaofeng},
  journal={IEEE Transactions on Power Systems},
  volume={29},
  number={2},
  pages={656--665},
  year={2013},
  publisher={IEEE}
}

@article{rogers2020,
  title={Probabilistic modelling of wind turbine power curves with application of heteroscedastic Gaussian process regression},
  author={Rogers, TJ and Gardner, P and Dervilis, N and Worden, K and Maguire, AE and Papatheou, E and Cross, EJ},
  journal={Renewable Energy},
  volume={148},
  pages={1124--1136},
  year={2020},
  publisher={Elsevier}
}

@article{pandit2020,
  title={Gaussian process power curve models incorporating wind turbine operational variables},
  author={Pandit, Ravi Kumar and Infield, David and Kolios, Athanasios},
  journal={Energy Reports},
  volume={6},
  pages={1658--1669},
  year={2020},
  publisher={Elsevier}
}

@article{2026probabilistic,
  title={Probabilistic wind power modelling via heteroscedastic non-stationary gaussian processes},
  author={Ladopoulou, Domniki and Hong, Dat Minh and Dellaportas, Petros},
  journal={Energy Reports},
  volume={15},
  pages={108895},
  year={2026},
  publisher={Elsevier}
}

@article{Fiocchi_2025,
  title={Probabilistic multilayer perceptrons for wind farm condition monitoring},
  author={Fiocchi, Filippo and Ladopoulou, Domniki and Dellaportas, Petros},
  journal={Wind Energy},
  volume={28},
  number={4},
  pages={e70012},
  year={2025},
  publisher={Wiley Online Library}
}

\begin{appendices}
\section{Gaussian Process Regression}
\label{subsec:gp}

A GP $f: \mathbb{R}^D \rightarrow \mathbb{R}$ is a stochastic process such that, for any finite collection of inputs $X = \{x_i\}_{i=1}^N \subset \mathbb{R}^D$, the vector of function evaluations 
$f(X) = [f(x_1), \ldots, f(x_N)]^T$
follows a multivariate normal distribution \cite{Rasmussen2006GaussianLearning}. GPs satisfy the Kolmogorov extension theorem, which guarantees that all finite-dimensional marginals are mutually consistent under permutation and marginalisation \cite{kolmogorov1950}. This provides a principled Bayesian framework for modelling distributions over functions, enabling exact inference with predictive uncertainty.  In our experiments presented in Section \ref{sec:results}, we use GP regression as a benchmark model, so we briefly outline its formulation here.

In the regression setting, we are given a data vector $y =\{y_i\}_{i=1}^N \in \mathbb{R}^N$ whose entries are noisy evaluations of some function $f(\cdot)$ on a collection of $D$-dimensional vectors $X = \{x_i\}_{i=1}^N \in \mathbb{R}^{N \times D}$. Each $y_i$ is assumed to be a noisy observation of $f(x_i)$, with independent Gaussian noise of mean $0$ and variance $\sigma^2$. Placing a GP prior over $f(\cdot)$, with mean function $\mu(\cdot)$ and covariance kernel $k_\theta(\cdot,\cdot)$, gives the joint distribution
\begin{align}
f(X) \sim \mathcal{N} (\mu(X),K(X,X)),
\end{align}
where $\mu(X) = [\mu(x_1),\ldots,\mu(x_N)]^T$ and $K(X,X)_{ij} = k_\theta(x_i,x_j)$ for all $i,j$.  

The predictive distribution for future observations $y^*$ with covariates $X^*$ is Gaussian with mean and variance
\begin{align}
E(y^* | y) &= \mu(X^*)+K(X^*,X) A^{-1} (y-\mu(X)), \\
V(y^* | y) &= K(X^*,X^*) -K(X^*,X) A^{-1} K(X,X^*).
\end{align} 
where $A=K(X,X)+\sigma^2 I_N$, with $\sigma^2$
denoting the observation noise variance and $I_N$
the $N\times N$ identity matrix.
For our experiments, we use the RBF kernel:
\begin{equation}
    k_{\text{RBF}}(x_i,x_j | \theta) = \sigma_f^2 \exp\left(-\frac{1}{2} \sum_{d=1}^D \frac{(x_{i,d} - x_{j,d})^2}{\ell_d^2}\right),
\end{equation} 
where $\theta=\{\sigma^2_f,\ell_1,\ell_2,\dots, \ell_D \}$ are the kernel hyperparameters, with $\ell_d$ denoting the length-scale parameters and $\sigma_f^2$ representing the output variance.  

While the RBF kernel is widely used and often effective in practice, encoding realistic prior assumptions through kernel design remains challenging, particularly for larger datasets and in higher-dimensional spaces. In addition, GPs assume Gaussian finite-dimensional marginals, restricting the class of functions they can represent. These issues motivate the exploration of alternative stochastic process models that relax Gaussianity while retaining uncertainty quantification and other desirable characteristics such as exchangeability.

\section{Diffusion Probabilistic Models}
\label{subsec:dpm}
Diffusion probabilistic models (DPMs) were introduced as a class of generative models inspired by non-equilibrium thermodynamics \cite{sohldickstein2015deepunsupervisedlearningusing}. The central idea is to define a forward process, in which the structure of the data distribution is gradually destroyed by the iterative addition of Gaussian noise, and a corresponding reverse process that learns to invert this corruption.

The denoising diffusion probabilistic model (DDPM) \cite{ho2020denoisingdiffusionprobabilisticmodels} simplified and stabilised training and has become the standard formulation. Following \cite{ho2020denoisingdiffusionprobabilisticmodels, dutordoir2023neural}, the forward process begins from the input data distribution $q(s_0)$, where $s_0 \in \mathbb{R}^D$ denotes a $D$-dimensional data vector.  It defines a fixed Markov chain over the sequence $s_{0:T} = \{s_0,\ldots,s_T\}$ with density
\begin{equation}
q(s_{0:T}) = q(s_0) \prod_{t=1}^T q(s_t \mid s_{t-1})
\end{equation}
and conditional distributions
\begin{equation}
q(s_t \mid s_{t-1}) = \mathcal{N}\!\left(\sqrt{1-\beta_t}\,s_{t-1}, \, \beta_t I_D\right),
\label{eq:ddpm_f}
\end{equation}
where $\{\beta_t \in (0,1)\}_{t=1}^T$ is a variance schedule controlling the magnitude of injected noise and $I_D$ is the $D \times D$ identity matrix. After $T$ steps, the distribution approaches Gaussian noise, $q(s_T) \approx \mathcal{N}(0,I)$. 
Early implementations used a linear variance schedule, while later work 
demonstrated that a cosine schedule improves performance by distributing noise 
more evenly across diffusion steps~\cite{nichol2021improveddenoisingdiffusionprobabilistic}. 
A variance schedule defines the cumulative noise parameter
\begin{equation}
\bar{\alpha}_t = \prod_{j=1}^t (1 - \beta_j),
\end{equation}
which quantifies the retained signal at step $t$. In the cosine schedule,
\begin{equation}
\bar{\alpha}_t = \frac{f(t/T)}{f(0)}, 
\qquad f(\lambda) = \cos^2\!\left(\tfrac{\pi}{2}\,\tfrac{\lambda+s}{1+s}\right),
\label{eq:cosine_schedule}
\end{equation}
where $T$ is the total number of steps and $s$ a small offset. This design 
avoids excessive noise at the beginning and end of the process, yielding more 
stable training and improved sample quality.

The reverse process is intractable in closed form, but can be approximated by an NN that predicts the injected noise. The mean of the reverse transition is parameterised as \begin{equation}
\mu_\theta(s_t, t) = \frac{1}{\sqrt{\alpha_t}} \Big( s_t - \frac{\beta_t}{\sqrt{1-\bar{\alpha}_t}} \, \epsilon_\theta(s_t, t) \Big),
\label{eq:ddpm_b} \end{equation} where $\alpha_t = 1-\beta_t$, $\bar{\alpha}_t = \prod_{j=1}^t \alpha_j$, and $\epsilon_\theta : \mathbb{R}^D \times \{1,\dots,T\} \to \mathbb{R}^D$ is an NN that predicts the injected noise vector. Training is carried out by minimising the score-matching loss \begin{equation}
    \mathbb{E}_{t, s_0, \epsilon}\big[ \lVert \epsilon - \epsilon_\theta(s_t, t) \rVert^2 \big],
\label{eq:ddpm_loss}
\end{equation}
where $s_t = \sqrt{\bar{\alpha}_t} s_0 + \sqrt{1-\bar{\alpha}_t} \epsilon$ and $\epsilon \sim \mathcal{N}(0,I)$.
Since $q(s_T) \approx \mathcal{N}(0,I)$, generating new samples requires only running the learned reverse process starting from Gaussian noise.

This framework enabled diffusion models to achieve state-of-the-art generative performance, rivalling generative adversarial networks (GANs) and autoregressive models. Building on DDPMs, RePaint \cite{lugmayr2022RePaintinpaintingusingdenoising} adapted the approach to image inpainting by conditioning on context points during the reverse process, without altering the pre-trained DDPM. This method achieved high-quality and diverse reconstructions under challenging conditions, outperforming autoregressive and GAN-based baselines.  

In parallel, NPs were proposed as a stochastic process model combining NNs with an encoder–decoder architecture \cite{garnelo2018neuralprocesses}. NPs learn to map a set of context points into a latent task representation, which is then used by a decoder to generate predictions at target inputs. This provides scalability and the ability to adapt to new tasks with only a few observations, but standard NPs often underfit and produce poorly calibrated uncertainty \cite{kim2019attentiveneuralprocesses}. Latent and attentive variants improve correlations between targets, but still fall short of fully consistent Bayesian behaviour \cite{dutordoir2023neural}.  

NDPs unify these two developments, drawing on diffusion-based generative modelling to overcome the Gaussian assumptions of NPs while retaining their scalability and adaptability. NDPs extend diffusion models from distributions over data to distributions over functions \cite{dutordoir2023neural}, thereby enabling uncertainty-aware regression.

\section{Single-Point Diffusion Trajectories}
\label{app:single_point_diffusion}

\begin{figure}[!ht]
\centering
\begin{subfigure}[b]{1\textwidth}
\includegraphics[width=\textwidth]{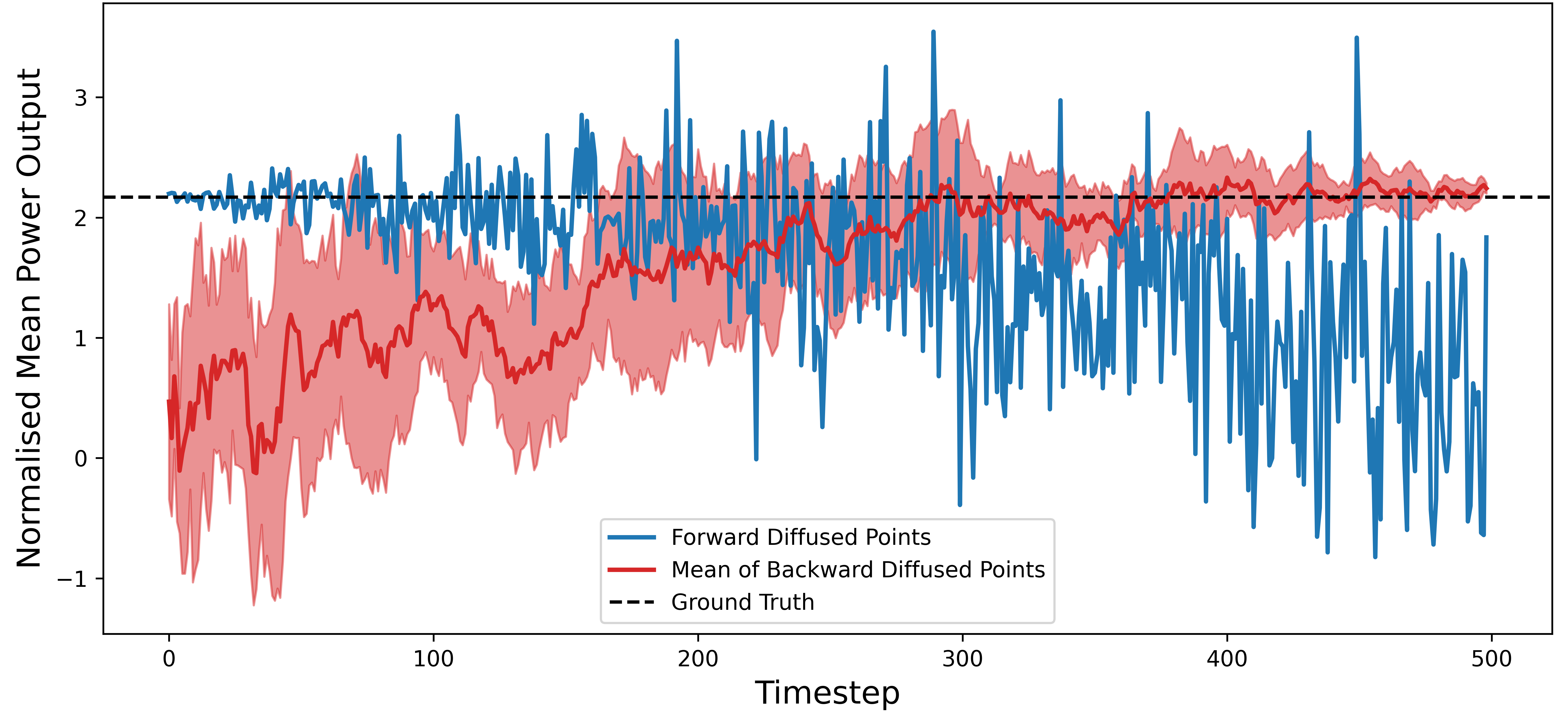}
\caption{Conditional sampling}
\label{fig:dd_fb_p_cond}
\end{subfigure}

\vspace{0.2cm}

\begin{subfigure}[b]{1\textwidth}
\includegraphics[width=\textwidth]{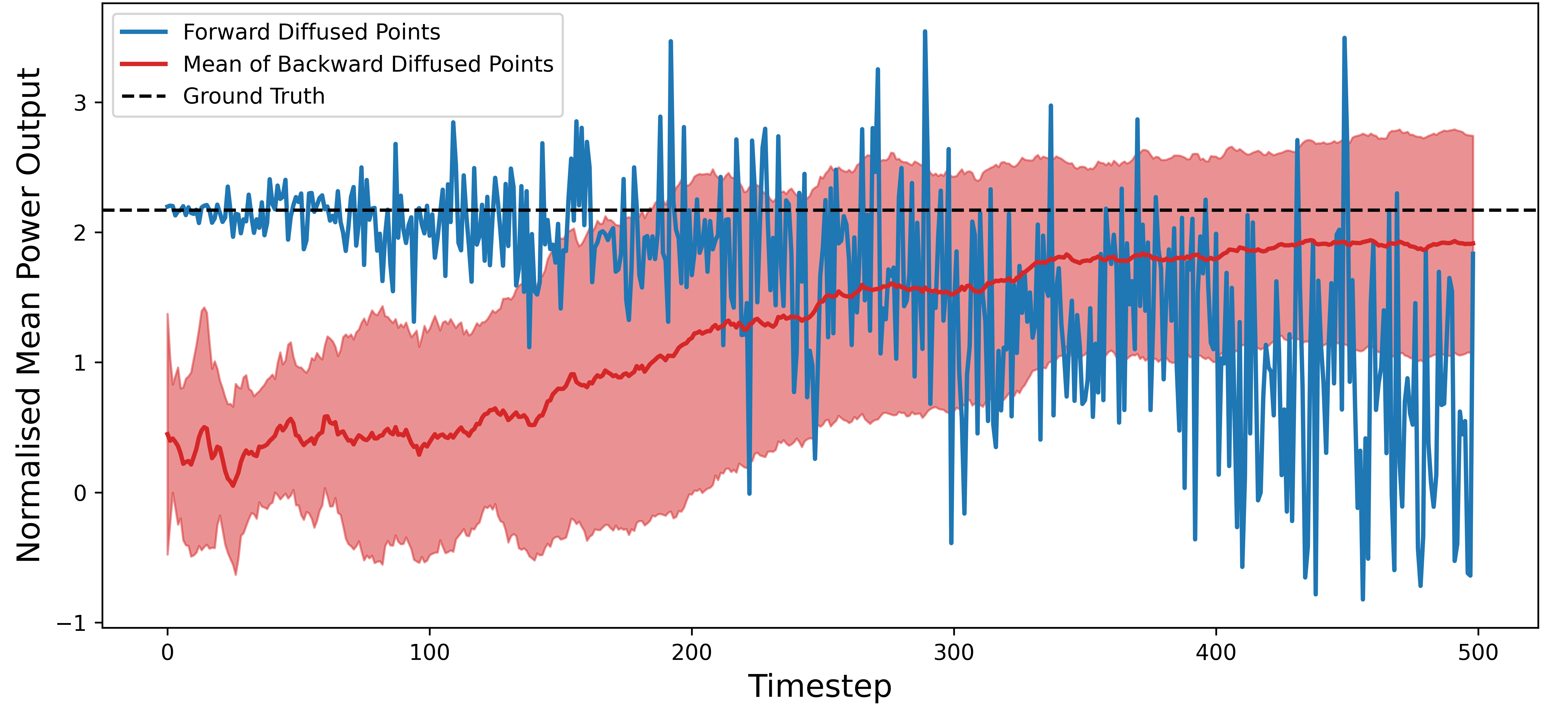}
\caption{Unconditional sampling}
\label{fig:dd_fb_p_uncond}
\end{subfigure}

\caption{Forward and reverse trajectories for a single input point under (a) conditional and (b) unconditional sampling. In the forward process (blue), Gaussian noise is gradually added. In the reverse process (red), noise is progressively removed. The shaded area shows one standard deviation.}
\label{fig:dd_fb_p}
\end{figure}

To illustrate the dynamics more clearly, Fig.~\ref{fig:dd_fb_p} depicts the forward and reverse trajectories for a single input point. In the forward process (blue), Gaussian noise gradually corrupts the true value. In the reverse process (red), starting from noise, the trajectories progressively recover the learned data structure. Conditional sampling is guided by the available observations, whereas unconditional sampling generates values from the learned prior without being constrained to recover the specific true observation.
\section{Computational Complexity Analysis}
\label{sec:computational_complexity}

The computational cost of MT-NDPs arises from two main
components: the task encoder and the diffusion denoising
network. Let $C$ and $N$ denote the numbers of context
and target observations, respectively, $T$ the number
of diffusion steps, $S$ the number of predictive samples,
and $B$ the training batch size. The task encoder processes the $C$ context observations
using a multilayer perceptron, followed by mean
aggregation. Its computational cost is
$\mathcal{O}(Cd_E)$, where $d_E$ denotes the cost of
encoding one context observation. Therefore, for a
fixed encoder architecture, the encoding cost increases
linearly with context size. During training, a diffusion timestep is sampled
randomly, so the denoising network is evaluated only
once per training example rather than over the entire
diffusion trajectory. The approximate computational
cost of one training iteration is therefore

\begin{equation}
{\mathcal{O}\left(
B\left(Cd_E+C_\theta(N)\right)
\right),
}
\end{equation}
where $C_\theta(N)$ denotes the computational cost
of one denoising-network evaluation on $N$ observations.
This cost depends on the network architecture and can
increase quadratically with the number of observations
when full self-attention is used.

At inference time, the task encoder constructs a
representation from the available context observations.
Generating $S$ predictive samples then requires $T$
reverse diffusion steps per sample. Assuming that
the task representation is computed once and reused
during sampling, the computational complexity is
approximately

\begin{equation}
\mathcal{O}\left(
Cd_E+STC_\theta(N)
\right).
\label{eq:mtndp_complexity}
\end{equation}

When RePaint-based conditional sampling is used,
noisy context observations are combined with the
target observations during the reverse diffusion
process. Consequently, the denoising network operates
on $C+N$ observations. RePaint may also introduce
additional denoising evaluations through repeated
forward and reverse steps. Its approximate
computational complexity is therefore

\begin{equation}
\mathcal{O}\left(
Cd_E+STRC_\theta(C+N)
\right),
\label{eq:mtndp_RePaint_complexity}
\end{equation}
where $R$ represents the average number of denoising
evaluations per diffusion timestep. These expressions show that the computational cost
of MT-NDP inference increases with the number of
diffusion steps and predictive samples. RePaint
introduces additional computational overhead because
the denoising network processes both context and
target observations and may require additional
denoising evaluations. In comparison, neural-process models generally
produce predictions without iterative diffusion
sampling, resulting in lower inference costs.
Exact Gaussian processes typically require
$\mathcal{O}(n^3)$ computational operations and
$\mathcal{O}(n^2)$ memory for training with $n$
observations, although approximations can reduce
these costs. Therefore, MT-NDPs do not necessarily provide
computational advantages over alternative
probabilistic regression methods. Instead, their
main benefit is the flexibility to model
task-dependent predictive distributions through
diffusion-based sampling.

\section{Wind Farm Dataset and Preprocessing} 
\label{dataset}

\begin{figure}[!ht]
    \centering
    \begin{subfigure}[b]{\textwidth}
        \includegraphics[width=0.95\textwidth]{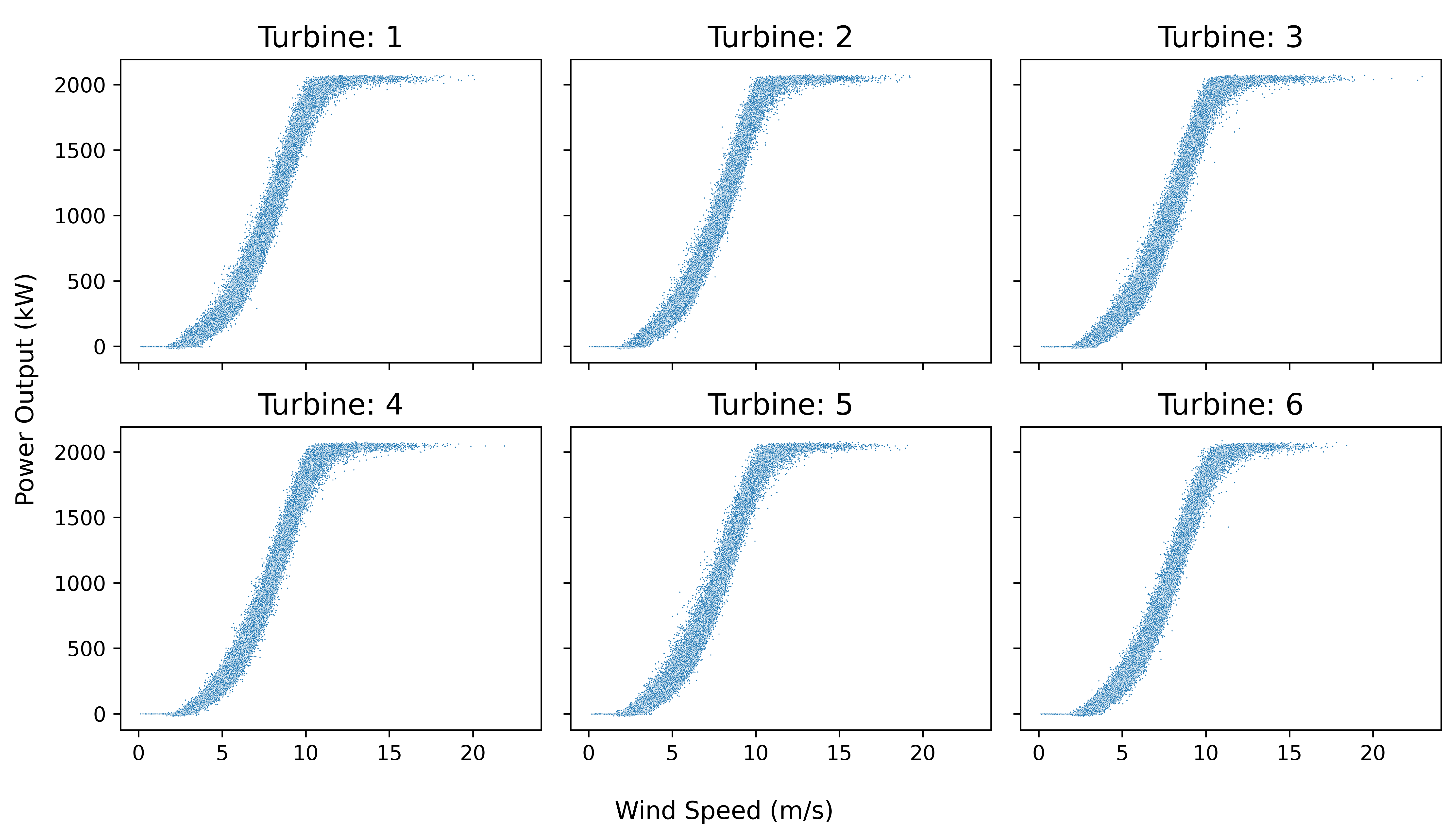}
        \caption{}
        \label{fig:scada}
    \end{subfigure}
    \hfill
    \begin{subfigure}[b]{\textwidth}
        \includegraphics[width=0.95\textwidth]{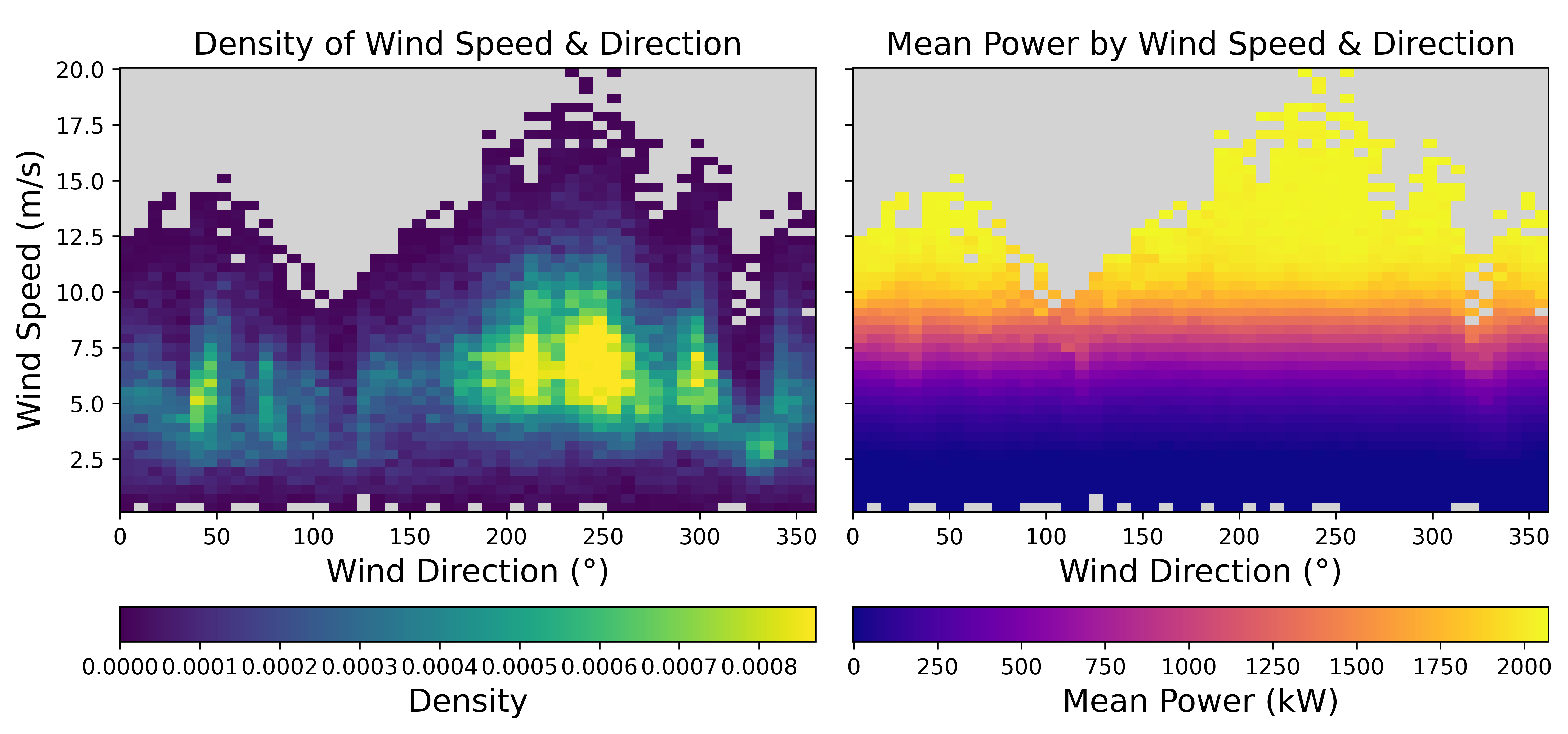}
        \caption{}
        \label{fig:eda}
    \end{subfigure}
\caption{SCADA data for the six turbines from the Kelmarsh wind farm \cite{Plumley2022} after removing standbys and warnings using the operational status and events file. (a) Wind power curves showing mean power output (kW) by mean wind speed (m/s).  
(b) Joint distribution of mean wind speed (m/s) and mean wind direction (\textdegree) for turbine~1 (left), and mean power output (kW) by mean wind speed (m/s) and wind direction (\textdegree) for turbine~1 (right).}
    \label{fig:combined1}
\end{figure}

An operational status and events file was used to filter the SCADA data and ensure consistent modelling of normal turbine behaviour. These logs provide detailed information on turbine operating states, including technical failures as well as operational or environmental standbys and warnings. To enable accurate modelling, all data associated with out-of-control conditions were removed so that model training was based solely on stable operating periods. Specifically, records corresponding to standbys, warnings, and operational stops were excluded. In addition, data from the week preceding each forced outage was discarded to reduce the likelihood of capturing deteriorating states. The resulting data elimination process is also described and illustrated in \cite{Fiocchi_2025}.

\begin{figure}[!ht]
    \centering
    \begin{subfigure}[b]{\textwidth}
        \includegraphics[width=0.95\textwidth]{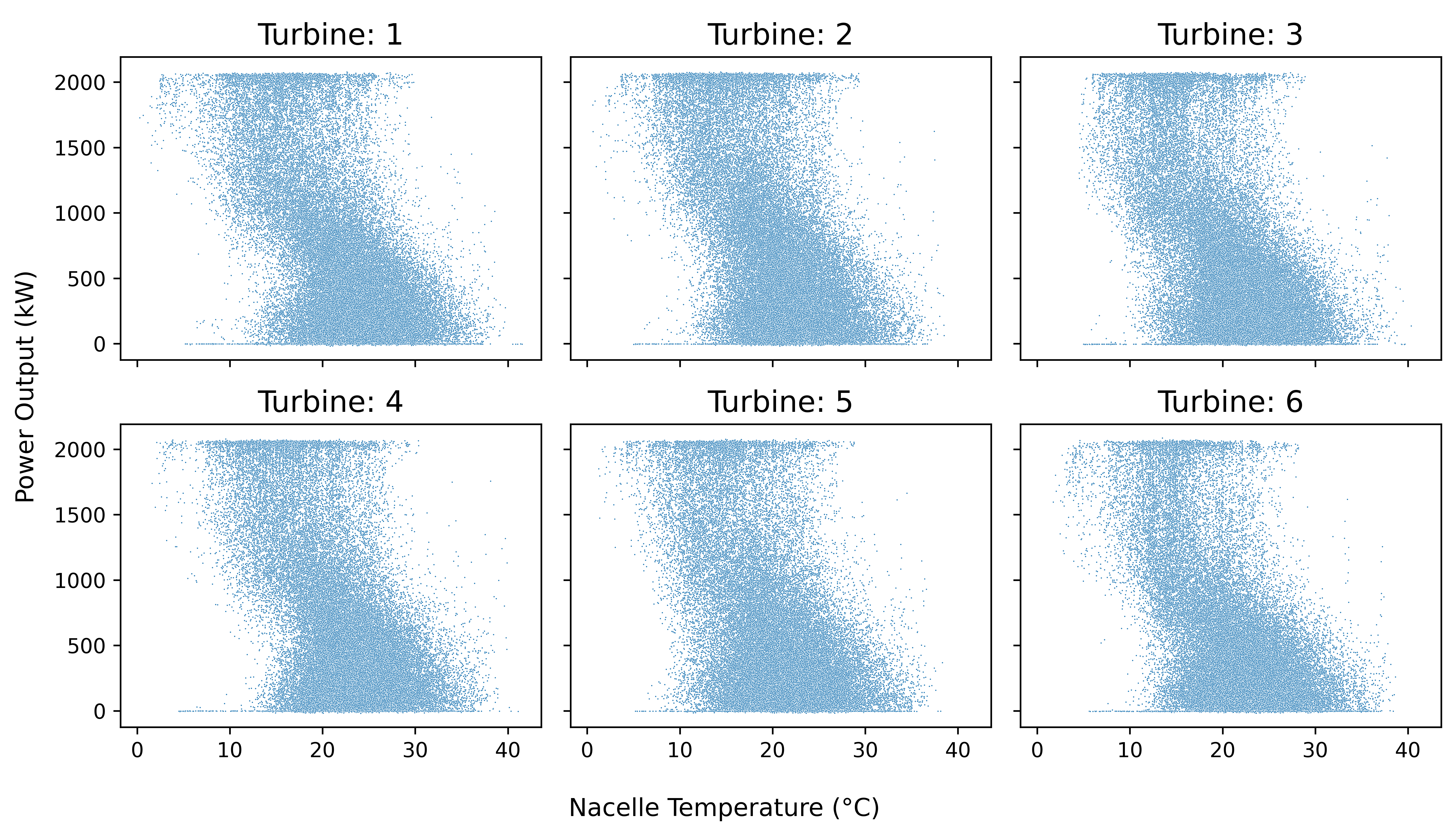}
        \caption{}
        \label{fig:eda_nt_power_curves}
    \end{subfigure}
    \hfill
        \begin{subfigure}[b]{\textwidth}
        \includegraphics[width=0.95\textwidth]{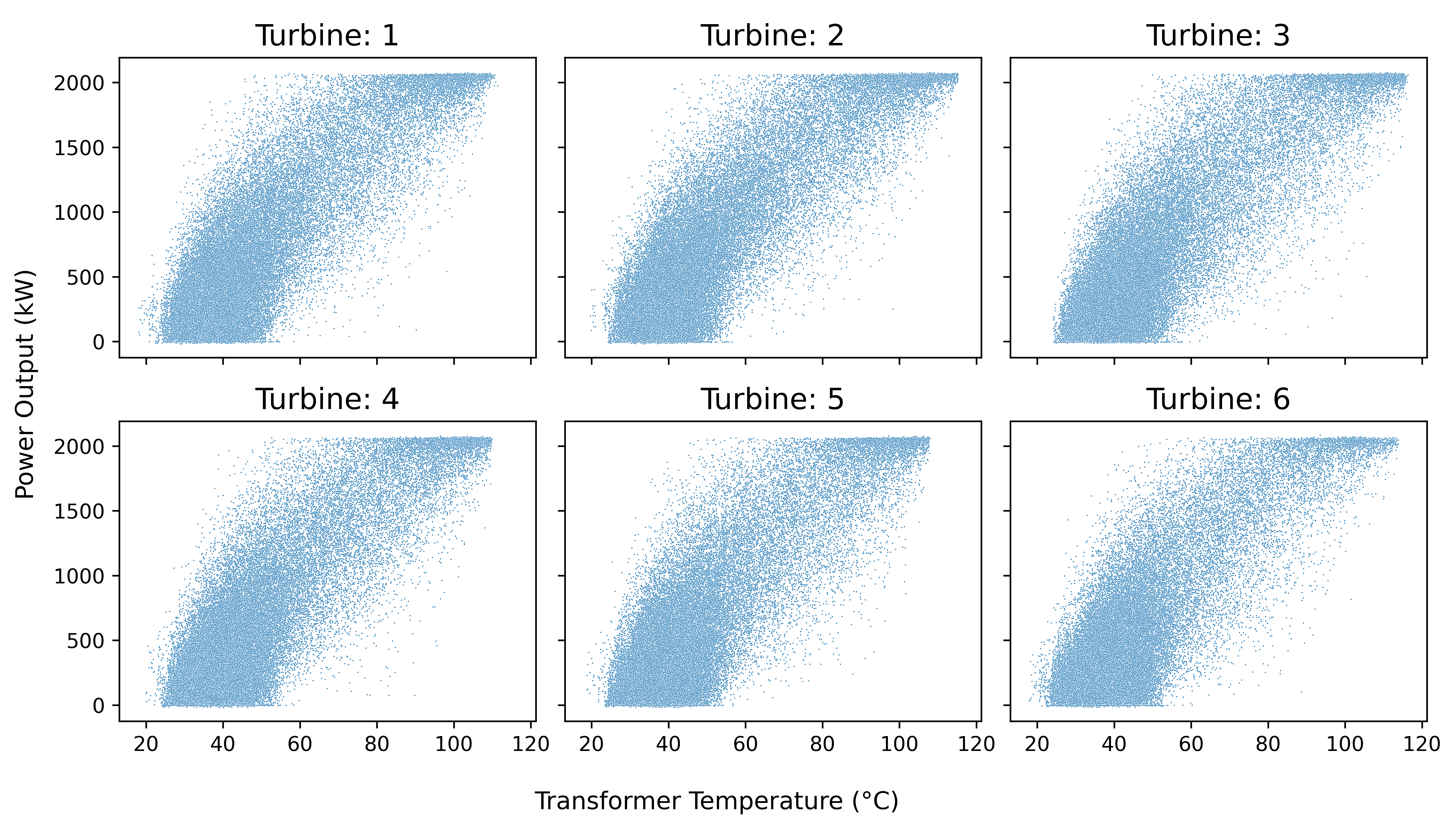}
        \caption{}
        \label{fig:eda_tt_power_curves}
    \end{subfigure}
    \caption{SCADA data for the six turbines from the Kelmarsh wind farm \cite{Plumley2022} after removing standbys and warnings using the operational status and events file.  (a) Mean power output (kW) by mean nacelle temperature (\textdegree C). (b) Mean power output (kW) by mean transformer temperature (\textdegree C).}
    \label{fig:combined2}
\end{figure}

Fig.~\ref{fig:scada} shows the wind speed–power curves for the six turbines after filtering. The curves are highly consistent and non-linear across wind turbines, with relatively few points observed at very high and very low wind speeds. These sparse regions are particularly challenging for probabilistic models such as GPs (see Section~\ref{sec:results}, Fig.~\ref{fig:1d_gp_ndp_fit}).  To examine directional effects, Fig.~\ref{fig:eda} presents the joint distribution of wind speed, wind direction, and power output for wind turbine~1. As expected, wind speed is the dominant component of variability. However, even at fixed wind speeds, power output varies with wind direction, likely reflecting terrain influences.  Fig.~\ref{fig:eda_nt_power_curves} presents mean nacelle temperature against mean power output. The pattern is more complex: rather than a monotonic trend, the relationship is approximately S-shaped. At intermediate power output, around 1000 kW, nacelle temperatures are on average lower than at low and high outputs. This non-linear behaviour highlights nacelle temperature as a particularly interesting covariate, since it reflects both operational load and thermal dynamics within the wind turbine.  In Fig.~\ref{fig:eda_tt_power_curves}, we show the relationship between mean transformer temperature and mean power output across the six wind turbines. The curves are highly consistent, suggesting that transformer temperature scales predictably with power output and may provide additional explanatory power beyond wind speed.

\section{Baseline Implementation Details} \label{app:baseline_implementation}

The MT-NP baseline uses a shared latent neural process with a permutation-invariant context encoder and a Gaussian decoder. The encoder and decoder employ multilayer perceptrons with hidden dimension $128$, and the latent representation has dimension $32$. The ANP baseline extends this architecture with a deterministic cross-attention mechanism using $8$ attention heads, alongside a latent representation of dimension $32$. Both neural-process baselines are trained for $12{,}000$ optimisation steps using Adam with a learning rate of $3\times10^{-4}$ and batch size $32$. Their training objectives combine Gaussian negative log-likelihood with a Kullback--Leibler divergence term weighted by $0.01$. The MT-GP baseline uses an intrinsic coregionalisation model with a rank-$2$ task covariance matrix and a five-dimensional radial basis function kernel with automatic relevance determination. A Gaussian likelihood is employed, and model parameters are optimised by maximising the exact marginal likelihood using Adam with a learning rate of $0.08$ for $75$ iterations.  To ensure comparability, all models are evaluated using identical held-out test observations and context sets, with predictive performance assessed using the same accuracy and calibration metrics. The neural-process baselines are trained jointly across the training tasks, while the MT-GP incorporates the available context observations from the held-out task during task-specific fitting. 

\section{Supplementary Wind Energy Experiments}\label{sec:additional_wind}
We first compare a one-dimensional GP with a one-dimensional NDP, using only wind speed as a feature. In this setting, wind turbines are nearly indistinguishable, consistent with exploratory analysis (Appendix, Fig.~\ref{fig:scada}). For the GP, we mimic the NDP setup, training on turbines $2$–$6$ and testing on turbine $1$. Due to scalability limits, the GP is trained on a subsample of $2000$ points evenly drawn from the five turbines. We compare this GP baseline only to the NDP with no context points, ensuring a fair assessment of prior quality.

This comparison indicates that NDPs can be competitive with GPs on this dataset, motivating their use as a foundation for more expressive models. We then investigate the effect of increasing input dimensionality by comparing one-dimensional and five-dimensional NDPs. The higher-dimensional model additionally incorporates wind direction (encoded via sine and cosine components), nacelle temperature, and transformer temperature. These variables were selected based on exploratory analysis (Fig.~\ref{fig:combined2}), which revealed strong and nonlinear relationships with power output. 

\subsection{Comparison of Gaussian Processes and NDPs}\label{res:gpvsndp}
We first compare the one-dimensional NDP without context points against a GP baseline. The effect of context on NDP performance is examined separately in Sections \ref{sec:res_mtndp} and Appendix Section \ref{res:1dvs5d}.
\begin{figure}[!ht]
    \begin{center}\includegraphics[width=1\textwidth]{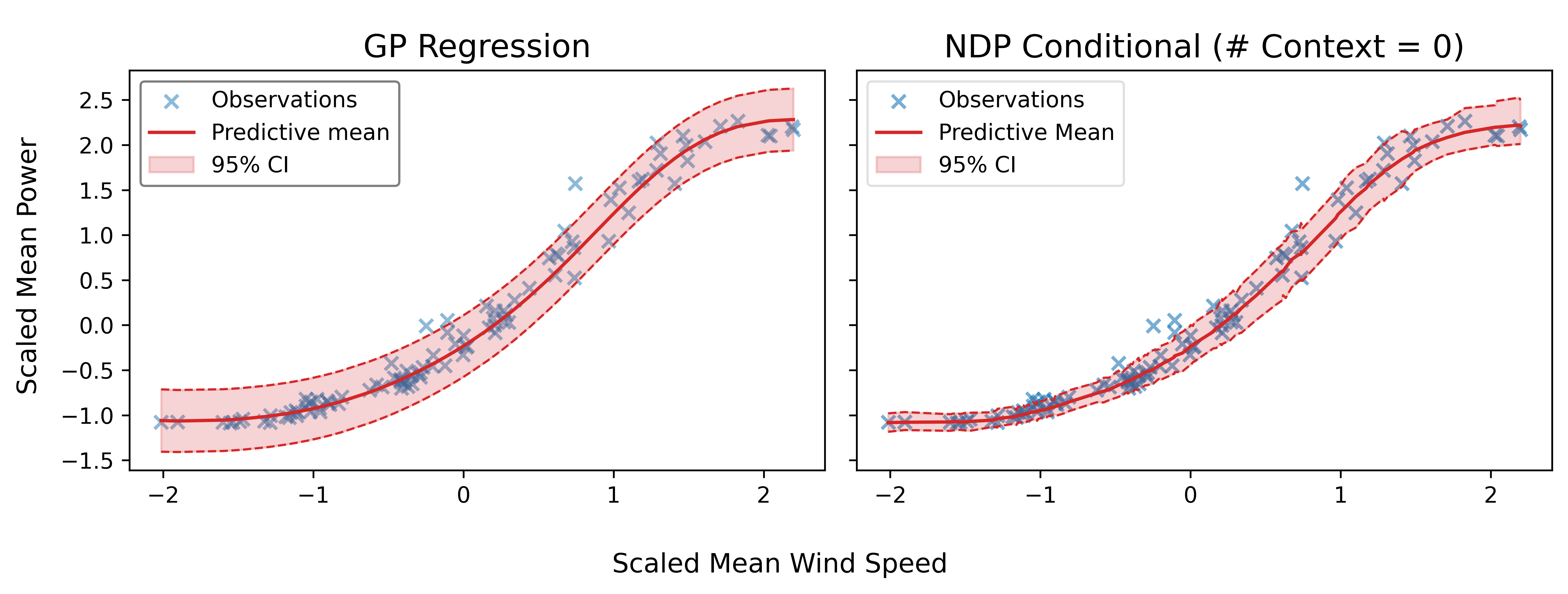}
        \caption{Fit of Gaussian process (GP) (left) and neural diffusion process (NDP) (right) on a test sample of $100$ points from turbine $1$, with $95$\% confidence intervals (CI) on normalised wind speed and power output.  1D: One-dimensional.
        \label{fig:1d_gp_ndp_fit}}
    \end{center}
\end{figure} Fig.~\ref{fig:1d_gp_ndp_fit} shows the fit of both models on a representative test function of $100$ points from turbine $1$. While both capture the mean trend well, their uncertainty estimates differ substantially. The NDP derives uncertainty from quantiles of reverse-diffusion trajectories, without assuming Gaussianity, whereas the GP relies on Gaussian predictive distributions. As a result, the GP is clearly underconfident, even in dense regions ($x \in [-1,1]$). This discrepancy becomes more pronounced in extrapolation regions ($x < -1$), where the NDP produces appropriately narrower confidence intervals. At the upper end of the domain ($x > 1$), improvements are less marked, though the NDP still provides slightly tighter uncertainty bounds than the GP.

\begin{figure}[!ht]
    \centering
    \begin{subfigure}[b]{0.57\textwidth}
        \includegraphics[width=\textwidth]{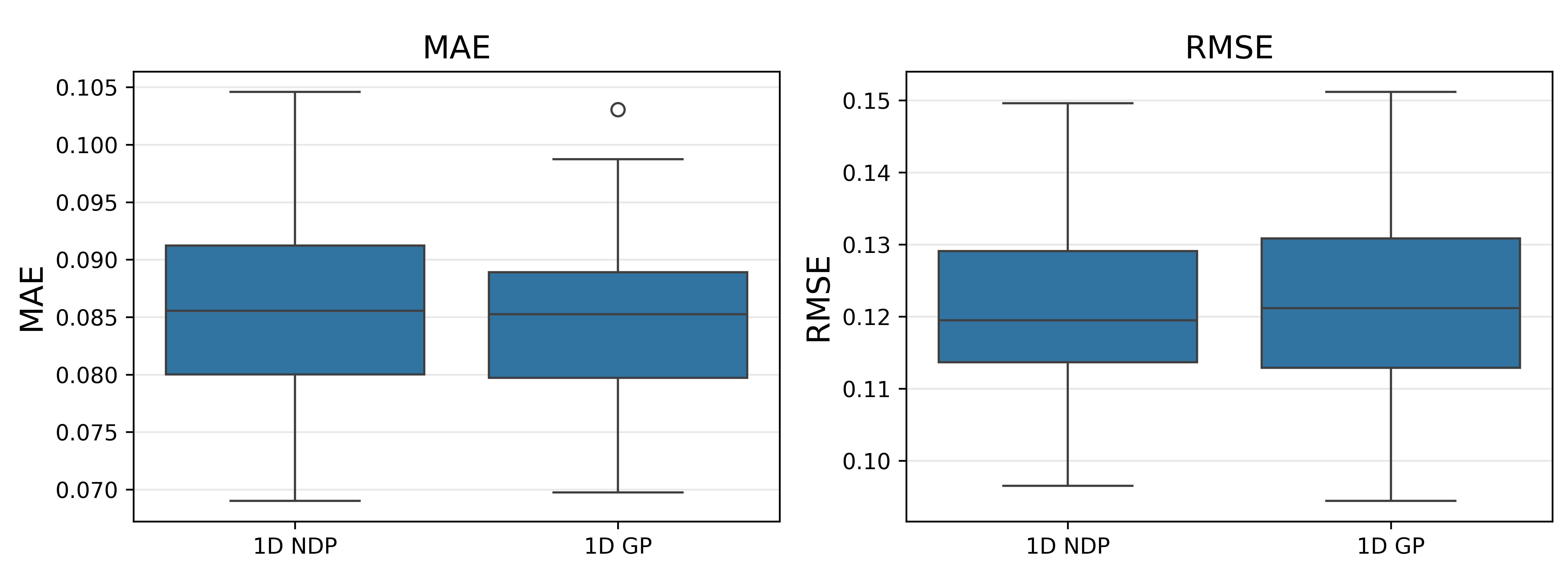}
        \caption{Mean absolute error (MAE, left) and root mean squared error (RMSE, right) distributions for Gaussian process (GP) and neural diffusion process (NDP) over $30$ test samples, each with $100$ points.}
        \label{fig:1d_loss}
    \end{subfigure}
    \hfill
    \begin{subfigure}[b]{0.42\textwidth}
        \includegraphics[width=\textwidth]{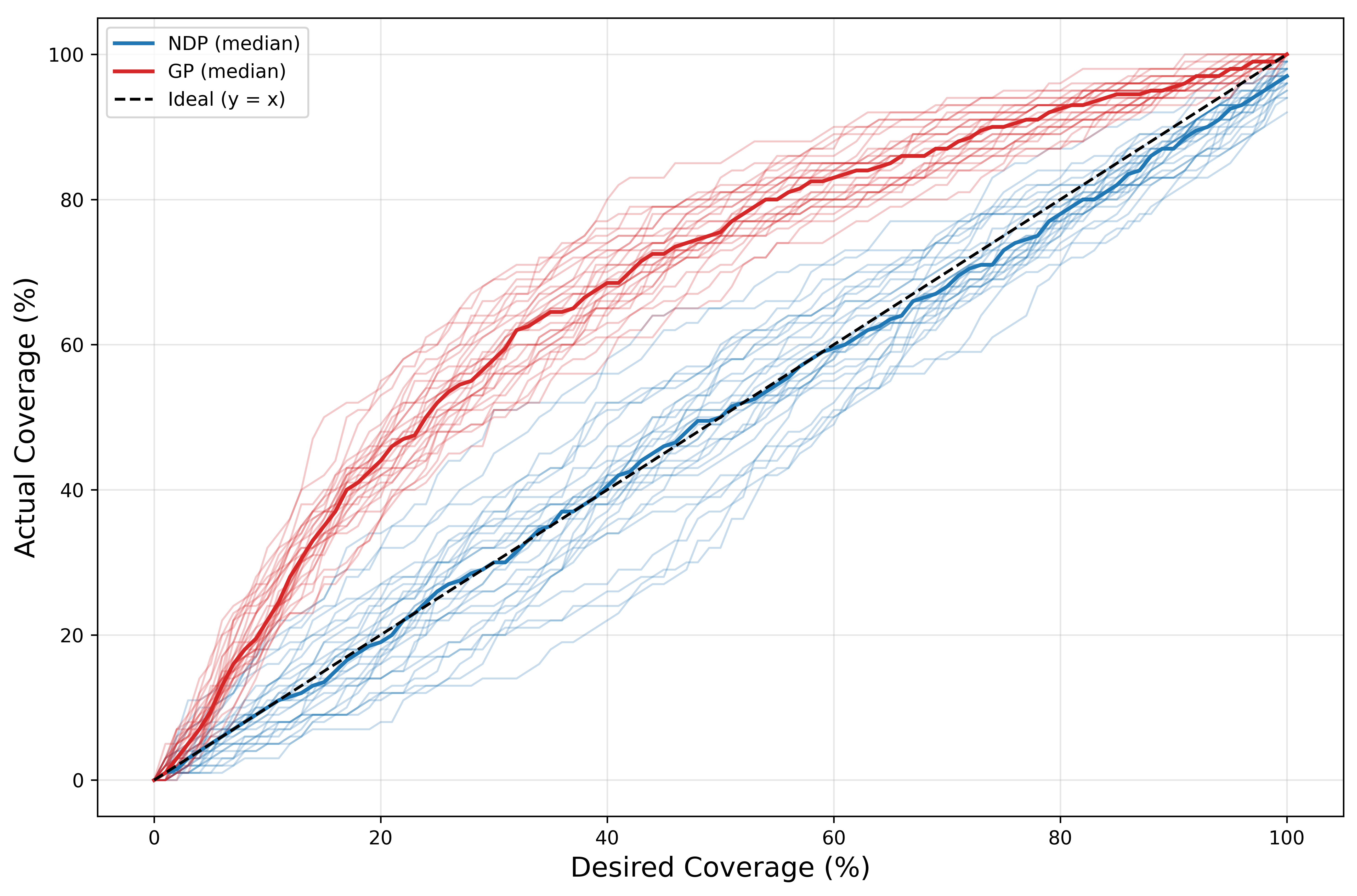}
        \caption{Observed vs.\ theoretical coverage probabilities for Gaussian process (GP) and neural diffusion process (NDP) over $30$ test samples, each with $100$ points.}
        \label{fig:1d_cov}
    \end{subfigure}
    \caption{Comparison of Gaussian process (GP) and neural diffusion process (NDP) on turbine $1$ test data. (a) Point prediction accuracy (MAE and RMSE distributions). (b) Predictive uncertainty calibration (coverage probabilities). 1D: One-dimensional, ST: single-task}
    \label{fig:1d_combined}
\end{figure}

Considering all $30$ test samples of $100$ points each, both models achieve comparable point prediction accuracy. As shown in Fig.~\ref{fig:1d_loss}, the distributions of MAE and RMSE for the GP and NDP are nearly indistinguishable, indicating that both capture the mean power–wind speed relationship effectively. The distinction between the models becomes clear when evaluating predictive uncertainty. Fig.~\ref{fig:1d_cov} shows the empirical coverage against the nominal coverage level. A perfectly calibrated model would follow the diagonal $y=x$ line. While the NDP is not flawless, it consistently lies closer to this ideal than the GP, which suffers from pronounced underconfidence across coverage levels. This highlights a key strength of NDPs: the ability to generate well-calibrated uncertainty estimates without relying on Gaussian assumptions.

\subsection{Effect of Input Dimensionality on NDP Performance}\label{res:1dvs5d}

The five-dimensional NDP extends the one-dimensional baseline by incorporating additional input features: the cosine and sine of wind direction, transformer temperature, and nacelle temperature, alongside wind speed.

\begin{figure}[!ht]
    \begin{center}
        
            \includegraphics[width=1\textwidth]{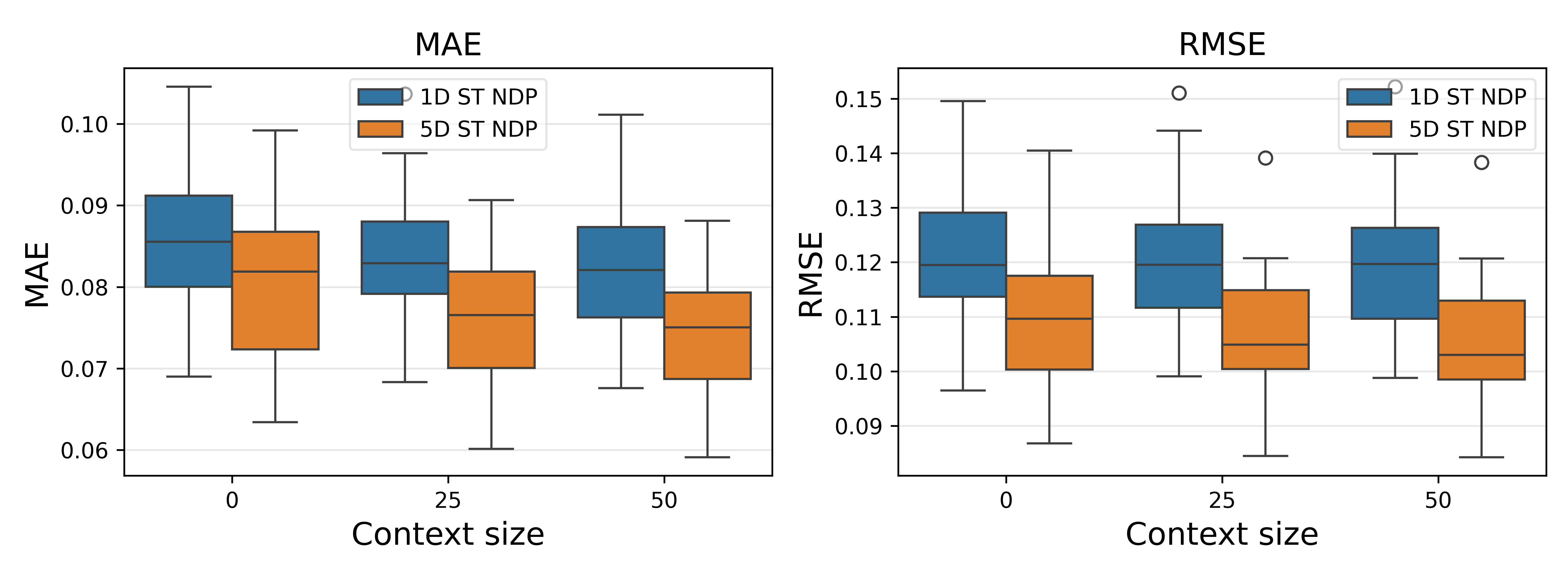}
            
        \caption{Distributions of mean absolute error (MAE, left) and root mean squared error (RMSE, right) for the one-dimensional (1D) and five-dimensional (5D) neural diffusion process (NDP) models, evaluated with $0$, $25$, and $50$ context points. Results are aggregated over $30$ test functions of $100$ points each from turbine~$1$. ST: single-task.
        \label{fig:1_5_loss}}
    \end{center}
\end{figure}

\begin{figure}[!ht]
    \begin{center}
            \includegraphics[width=1\textwidth]{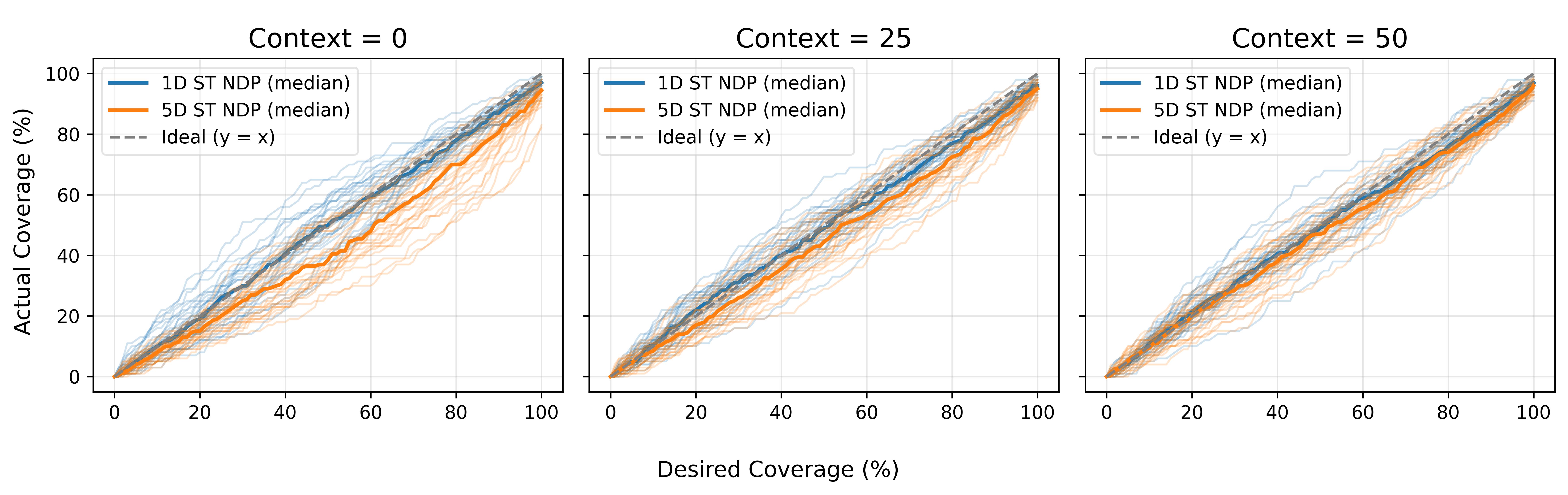}
\caption{Observed vs. theoretical coverage probabilities for the one-dimensional (1D) and five-dimensional (5D) neural diffusion process (NDP) models with $0$, $25$, and $50$ context points. Results are averaged over $30$ test functions of $100$ points each from turbine~$1$. ST: single-task.     \label{fig:1_5_cov}}
    \end{center}
\end{figure}

Fig.~\ref{fig:1_5_loss} compares the point prediction errors of the one-dimensional and five-dimensional NDP models across different context sizes. Incorporating more input features leads to a clear reduction in both MAE and RMSE, demonstrating the added explanatory power of the extended feature set. Notably, the benefit of context is much more apparent in the five-dimensional NDP model, where performance improves consistently as context size increases. In contrast, the one-dimensional NDP model shows little to no improvement when additional context points are provided.

\begin{figure}[!ht]
    \begin{center}
        \includegraphics[width=1\textwidth]{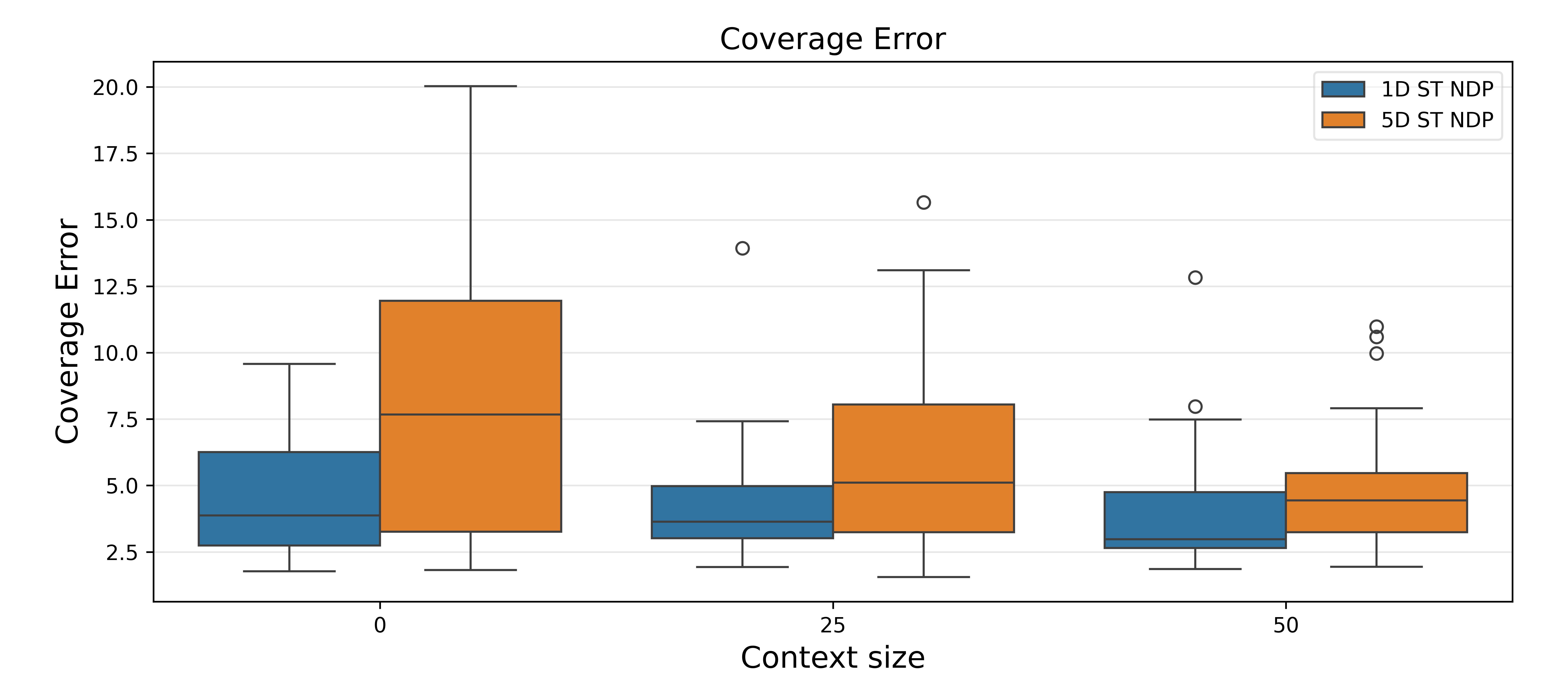}
\caption{Coverage error for the one-dimensional single-task neural diffusion process (1D ST NDP) and five-dimensional single-task neural diffusion process (5D ST NDP) models with $0$, $25$, and $50$ context points. Results are computed over $30$ test functions of $100$ points each from turbine~$1$.        \label{fig:1_5_cov_err}}
    \end{center}
\end{figure}
Fig.~\ref{fig:1_5_cov} presents observed versus nominal coverage for both models at different context sizes. The five-dimensional NDP shows steady improvement in calibration with increasing context, whereas the one-dimensional NDP remains largely unaffected by context size. Interestingly, the one-dimensional NDP achieves slightly better coverage alignment than the five-dimensional NDP when both are provided with $50$ context points. This raises the question of whether the five-dimensional NDP model would outperform the one-dimensional model given a larger context set, a question left open here due to computational constraints.

Finally, Fig.~\ref{fig:1_5_cov_err} summarises coverage performance using the CE metric. The results confirm that context improves calibration for the five-dimensional NDP, narrowing the gap to the one-dimensional NDP model. When considered alongside the substantial accuracy gains in Fig.~\ref{fig:1_5_loss}, the evidence suggests that the five-dimensional NDP provides a meaningful improvement over the one-dimensional NDP baseline, particularly in balancing accuracy with uncertainty calibration.

\begin{table}[t]
\centering
\caption{Out-of-sample performance of Gaussian process (GP) and single-task neural diffusion process (ST-NDP) models on turbine~$1$, averaged over $30$ test functions ($100$ points each). 
Metrics reported: mean absolute error (MAE, kW), root mean squared error (RMSE, kW), and coverage error (CE, \%). 1D: one-dimensional, 3D: three-dimensional, 5D: five-dimensional.}
\label{tbl:results_turbine1}
\begin{tabular}{lccc ccc ccc}
\toprule
 & \multicolumn{3}{c}{{Context = $0$}} 
 & \multicolumn{3}{c}{{Context = $25$}} 
 & \multicolumn{3}{c}{{Context = $50$}} \\
\cmidrule(lr){2-4} \cmidrule(lr){5-7} \cmidrule(lr){8-10}
\textit{Model} 
& MAE & RMSE & CE 
& MAE & RMSE & CE 
& MAE & RMSE & CE \\\midrule
1D GP      
& $51.636$ & $75.596$ & $18.075$ 
& -- & -- & -- 
& -- & -- & -- \\
1D ST-NDP  
& $52.603$ & $75.139$ & $4.584$ 
& $51.273$ & $74.568$ & $4.181$ 
& $50.545$ & $74.137$ & $3.938$ \\
3D ST-NDP  
& $60.967$ & $88.096$ & $9.731$ 
& $52.338$ & $74.312$ & $6.750$ 
& $50.855$ & $73.537$ & $5.369$ \\
5D ST-NDP  
& $49.429$ & $67.685$ & $8.351$ 
& $46.962$ & $66.007$ & $6.029$ 
& $45.791$ & $65.301$ & $4.816$ \\
\bottomrule
\end{tabular}
\end{table}

Table~\ref{tbl:results_turbine1} provides a quantitative summary of the comparisons, aggregating point prediction accuracy (MAE, RMSE) and predictive uncertainty calibration (CE) across all models and context sizes for turbine~$1$. The results confirm the visual trends: while the GP achieves similar MAE and RMSE to the NDP, it performs very poorly in terms of uncertainty calibration, with a coverage error more than four times higher. Among the NDP variants, incorporating additional features (from one-dimensional to five-dimensional) consistently reduces prediction error, while increasing the number of context points improves both accuracy and calibration.

\subsection{Supplementary Calibration Results}
\label{sec:wind_coverage}

\begin{figure}[!ht]
    \centering
    \includegraphics[width=\textwidth]{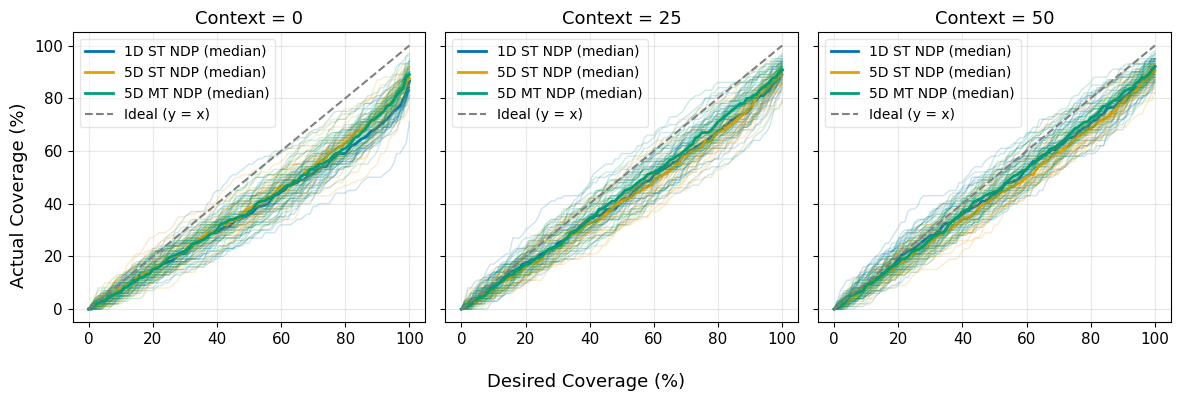}
    \caption{Empirical versus nominal coverage probabilities for the one-dimensional ST-NDP, five-dimensional ST-NDP, and five-dimensional MT-NDP on turbine~6 across context sizes $0$, $25$, and $50$.}
    \label{fig:paper_cov_t6}
\end{figure}

Fig.~\ref{fig:paper_cov_t6} presents the empirical versus nominal coverage probabilities for the three NDP variants across context sizes $0$, $25$, and $50$. These results complement the coverage error distributions reported in the main paper and provide a more detailed assessment of predictive uncertainty calibration.

\section{Supplementary Synthetic Ablation Results}
\label{app:synthetic_ablation}

\begin{figure}[!ht]
    \centering
    \includegraphics[width=\textwidth]{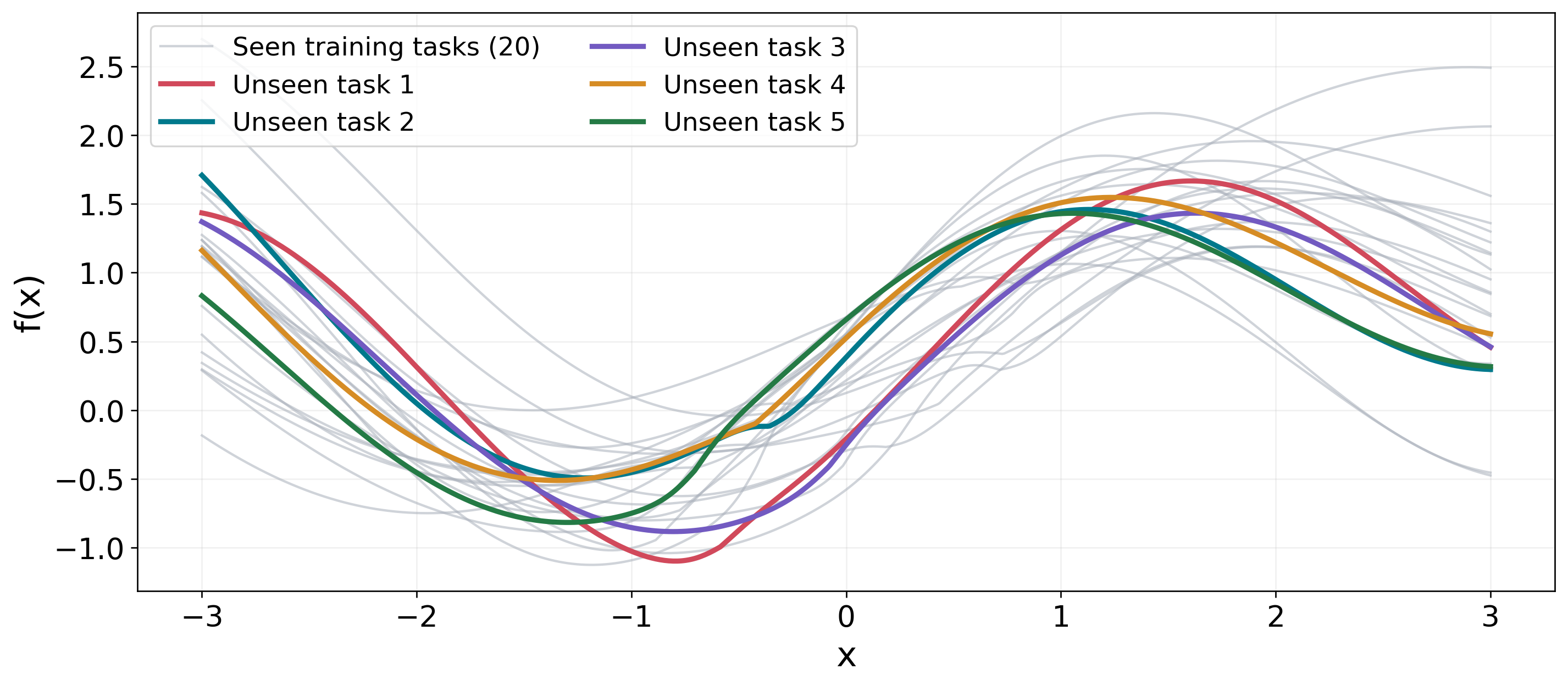}
    \caption{Synthetic multi-task regression dataset comprising 20 training tasks (grey) and five previously unseen test tasks (coloured). The functions are generated using a common data-generating process, with task-specific variation to represent differences between related regression tasks. The five held-out tasks are used to evaluate the generalisation performance of MT-NDP.}
    \label{fig:synthetic_example}
\end{figure}

To further investigate the effect of the task embedding dimension
and context size on the predictive performance of MT-NDP, we conduct
an ablation study using a synthetic multi-task regression dataset.
The dataset consists of 20 training tasks and five previously unseen
test tasks, as illustrated in Fig.~\ref{fig:synthetic_example}.
This setting allows us to evaluate the ability of MT-NDP to
generalise to new tasks while varying the amount of task-specific
context information available at prediction time. We consider three task embedding dimensions,
$\kappa \in \{4,8,16\}$, and four context sizes,
$C \in \{0,5,10,25\}$.
Predictive performance is assessed using MAE, RMSE and empirical coverage.
We also report the computational time required for inference.
Results are summarised in Table~\ref{tab:synthetic_ablation}
as mean (standard deviation) across five unseen tasks
and two repetitions.

\begin{table}[!ht]
\centering
\caption{Results of the synthetic ablation study evaluating the effect of task embedding dimension ($\kappa$) and context size ($C$) on the predictive performance of MT-NDP. The model is trained on
20 synthetic tasks and evaluated on five previously unseen tasks. Performance is assessed using mean absolute error (MAE), root mean squared error (RMSE), empirical 95\% coverage, and inference time in minutes. Results are reported as mean (standard deviation) across five test tasks and two repetitions.}
\label{tab:synthetic_ablation}
\small
\begin{tabular}{cccccc}
\toprule
$\kappa$ & $C$ & MAE & RMSE & 95\% Coverage (\%) & Time (min) \\
\midrule
4 & 0  & 0.433 (0.078) & 0.497 (0.094) & 67.8 (3.8) & 2.98 (0.12) \\
4 & 5  & 0.253 (0.078) & 0.309 (0.093) & 75.9 (4.0) & 8.25 (0.19) \\
4 & 10 & 0.194 (0.058) & 0.242 (0.074) & 82.0 (3.6) & 7.80 (0.20) \\
4 & 25 & 0.151 (0.034) & 0.190 (0.046) & 86.7 (3.2) & 8.31 (0.52) \\
\midrule
8 & 0  & 0.424 (0.093) & 0.491 (0.107) & 72.1 (3.9) & 3.86 (0.11) \\
8 & 5  & 0.259 (0.062) & 0.307 (0.069) & 79.9 (2.7) & 9.71 (0.26) \\
8 & 10 & 0.204 (0.045) & 0.250 (0.053) & 88.5 (3.6) & 9.45 (0.32) \\
8 & 25 & 0.140 (0.033) & 0.176 (0.046) & 92.2 (3.0) & 9.91 (0.15) \\
\midrule
16 & 0  & 0.421 (0.078) & 0.487 (0.090) & 72.9 (3.1) & 4.50 (0.13) \\
16 & 5  & 0.263 (0.090) & 0.323 (0.103) & 82.4 (4.1) & 10.97 (0.43) \\
16 & 10 & 0.199 (0.069) & 0.248 (0.084) & 88.7 (3.8) & 10.45 (0.37) \\
16 & 25 & 0.151 (0.036) & 0.191 (0.045) & 95.6 (3.5) & 10.69 (0.28) \\
\bottomrule
\end{tabular}
\end{table}

The results demonstrate that increasing the context size
generally improves predictive accuracy across all three
embedding dimensions. For example, with $\kappa=8$, the MAE
decreases from $0.424$ without context observations to $0.140$
with 25 context observations, while the RMSE decreases from
$0.491$ to $0.176$. This indicates that MT-NDP can effectively
incorporate task-specific observations to improve predictions
on previously unseen tasks. The effect of the task embedding dimension is less pronounced.
Although increasing $\kappa$ generally improves empirical
coverage, the lowest MAE and RMSE are obtained with
$\kappa=8$ and $C=25$. Increasing the embedding dimension
to $\kappa=16$ does not provide further improvements in
predictive accuracy at this context size, suggesting that larger
task embeddings are not necessarily beneficial in this setting. Finally, larger embedding dimensions are associated with
increased computational time. These findings suggest that
$\kappa=8$ provides a reasonable balance between predictive
accuracy, empirical coverage, and computational cost for
the synthetic experiment.

\section{Supplementary School Examination Experiments}
\label{app:schools}

To further evaluate MT-NDP under a different training
configuration, we conduct an additional leave-one-school-out
experiment using six schools. In each evaluation, five schools
are used for training and the remaining school is held out
for testing, with $C=20$ context observations.
Table~\ref{tab:schools_results2} reports the mean and
standard deviation across the six evaluations.

\begin{table}[!ht]
\centering
\caption{Predictive performance of MT-NDP on the School
Examination dataset under leave-one-school-out evaluation.
Six schools are considered, with five used for training
and the remaining school held out for testing in each
evaluation. Each held-out school is evaluated using
$C=20$ context observations. Results are reported as
mean (standard deviation) across the six evaluations.
MAE denotes mean absolute error, RMSE denotes root
mean squared error, Coverage 95\% refers to the empirical
coverage of nominal 95\% predictive intervals, and
Time (min) represents inference time in minutes per
held-out-school evaluation.}
\label{tab:schools_results2}
\small
\begin{tabular}{lcccc}
\toprule
Model & MAE & RMSE & Coverage 95\% (\%) & Time (min) \\
\midrule
MT-NDP & 0.805 (0.186) & 1.008 (0.190) &
95.148 (4.576) & 13.124 (0.459) \\
\bottomrule
\end{tabular}
\end{table}

\end{appendices}

\end{document}